\documentclass[journal]{IEEEtran}
\usepackage[T1]{fontenc}
\usepackage{bm}
\usepackage{tabularx}
\usepackage[table]{xcolor}
\usepackage{colortbl}
\usepackage{tablestyles}
\usepackage{cite} 
\usepackage{amssymb}
\usepackage{amsmath}
\usepackage{balance}  
\usepackage{graphicx} 
\usepackage{times}    
\usepackage{url}      
\usepackage[lofdepth,lotdepth]{subfig}  
\usepackage{multirow}
\usepackage{bbm}
\usepackage{pseudocode}
\usepackage{cite}
\usepackage{multirow}
\usepackage{bbm}
\usepackage{pseudocode}
\usepackage{color}
\usepackage{algpseudocode}
\usepackage[boxed,ruled]{algorithm2e}
\usepackage{comment}
\usepackage{color,soul}
\usepackage{lipsum}

\usepackage[table]{xcolor}
\usepackage{tablestyles}
\usepackage{tabu}
\usepackage{multirow}
\usepackage{colortbl}

\renewcommand{\hl}[1]{
#1}

\definecolor{kugray5}{RGB}{224,224,224}

\usepackage{booktabs}

\begin{document}

\title{From Personalized Medicine to Population Health: A Survey of mHealth Sensing Techniques}

\author{Zhiyuan Wang, Haoyi Xiong~\emph{Senior Member, IEEE}, Jie Zhang, Sijia Yang, Mehdi Boukhechba,\\ Laura E. Barnes, Daqing Zhang,~\emph{Fellow, IEEE}, and Dejing Dou~\emph{Senior Member, IEEE}
\thanks{Z. Wang, H. Xiong and D. Dou are with the Big Data Lab, Baidu Research, Baidu Inc., Haidian, Beijing, China.}
\thanks{Z. Wang, M. Boukhechba, and L.E Barnes are with the Department of Engineering Systems and Environment, University of Virginia, Charlottesville, Virginia, United States.}
\thanks{J. Zhang and D. Zhang are with the Department of Computer Science, Peking University, Haidian, Beijing, China.}
\thanks{S. Yang is with the School of Cyberspace Security, Beijing University of Posts and Telecommunications, Haidian, Beijing, China.}
\thanks{Corresponding Author: Haoyi Xiong, haoyi.xiong.fr@ieee.org}}

\maketitle
\IEEEpeerreviewmaketitle
\begin{abstract}
    Mobile Sensing Apps have been widely used as a practical approach to collect behavioral and health-related information from individuals and provide timely intervention to promote health and well-beings, such as mental health and chronic cares. As the objectives of mobile sensing could be either \emph{(a) personalized medicine for individuals} or \emph{(b) public health for populations}, in this work we review the design of these mobile sensing apps, and propose to categorize the design of these apps/systems in two paradigms -- \emph{(i) Personal Sensing} and \emph{(ii) Crowd Sensing} paradigms. While both sensing paradigms might incorporate with common ubiquitous sensing technologies, such as wearable sensors, mobility monitoring, mobile data offloading, and/or cloud-based data analytics to collect and process sensing data from individuals, we present two novel taxonomy systems from \emph{Sensing Objectives} and \emph{Sensing Paradigms} perspectives that can specify and classify apps/systems from aspects of the life-cycle of mHealth Sensing: \emph{(1) Sensing Task Creation \& Participation}, \emph{(2) Health Surveillance \& Data Collection}, and \emph{(3) Data Analysis \& Knowledge Discovery}. With respect to different goals of the two paradigms, this work systematically reviews this field, and summarizes the design of typical apps/systems in the view of the configurations and interactions between these two components. In addition to summarization, the proposed taxonomy system also helps figure out the potential directions of mobile sensing for health from both personalized medicines and population health perspectives. 
\end{abstract}


\IEEEpeerreviewmaketitle

\section{Introduction}

\hl{Mobile Sensing~\cite{lane2010survey} refers to a sensing paradigm leveraging ubiquitous sensors embedded in mobile devices (e.g., mobile phones, smartwatches) to monitor the environments, human behaviors, and interactions between human and environments in a human-centric manner\cite{puccinelli2005wireless,mohr2017personal}.} Lots of work studied the adoption of {mobile sensing techniques in health domains}~\cite{xiong2016sensus,perez2021wearables,yan2021towards} such as mental health~\cite{huang2016assessing} and chronic cares~\cite{sieverdes2013improving}. Early visionary works~\cite{choudhury2008mobile,lane2011bewell} proposed the basic framework of mobile health (mHealth) sensing techniques in nowadays that leverage ``non-invasive'' mobile sensing schemes~\cite{rabbi2011passive} to collect data for human activities recognition and infer the individual's health status using machine learning algorithms with longitude and real-time sensory data accordingly~\cite{huang2016assessing,sathyanarayana2016sleep,hung2016predicting,servia2017mobile}. 

\hl{Compared to traditional medical sensors that are frequently operated by health professionals to collect data from patients in clinic contexts, mHealth sensing relies on participation of voluntary users to obtain information for health-related well-beings in their daily life~\cite{amft2008recognition,rabbi2011passive,evans2016remote,basatneh2018health,nahum2018just}. Furthermore, the goal of mHealth sensing research is to study the innovative applications of mobile sensing techniques to collect behavioral data related to health and well-beings, while medical sensing aims at designing new measurement and instrument techniques for medical purposes~\cite{mosenia2017wearable}. More comparison between mHealth sensing and medical sensing could be found in Appendix.A. In this work, given the rapid development in such area, we propose to review and survey the recent progress in mHealth sensing techniques.}

\subsection{Health Issues and Health Outcomes of mHealth Sensing Apps/Systems}
\hl{There are several works reviewing and surveying the research problems \cite{steinhubl2013can,mukhopadhyay2014pervasive,varshney2014mobile,adibi2015mobile,silva2015mobile,pryss2019mobile}, emerging techniques~\cite{park2016emerging,boukhechba2020leveraging}, system design \cite{huang2012design,kelly2012investigation,almotiri2016mobile}, and prototyping tools \cite{lorenz2009mobile,sama2014evaluation,servia2017mobile} for mHealth Sensing apps/systems. In this work, we propose to first categorize the research works on mHealth sensing apps/systems with respect to the major health issues (e.g., depression and anxiety) that are covered by mHealth.}

\hl{Furthermore, for every major health issue reviewed here, we also discuss mHealth sensing research from the perspectives of \emph{personalized medicine} and \emph{population health} ---  two major health outcomes of modern healthcare\cite{chernichovsky2010integrating,carlsten2014genes,mega2014population}. Specifically, we would like to survey and compare the mHealth sensing techniques that handle to the health issues for either personalized medicine or population health purposes. In our work, we define these two outcomes as follows.}
\begin{itemize}
    \item \textbf{Personalized Medicine.} The personalized medicine focuses on individual patients--\emph{``with medical decisions, practices, interventions and/or products being tailored to the individual patient based on their predicted response or risk of disease''}~\cite{academy2015stratified}. Thus, the objective of personalized medicine is to improve and optimize the individual treatment effects through sensing, monitoring, and predicting their health status~\cite{pokorska2014personalized,amft2020personalized}.
    
    \item \textbf{Population Health.} The population health is defined as \emph{``the health outcomes~\cite{dewalt2004literacy} of a group of individuals, including the distribution of such outcomes within the group''}~\cite{kindig2003population}. The goal of population health is to promote the health of an entire human population~\cite{callahan1973definition}, where the approaches include discovering health outcomes, understanding patterns of health determinants, and policy making for interventions. 
    
\end{itemize}
\hl{Thus, the first part of this survey includes a comprehensive review on the research related to mHealth sensing apps/systems targeting at various health issues from the perspectives of population health and personalized medicine. In Appendix.B, we introduce the procedure that we collect and select the publications and health issues for review.}
%
%

\begin{figure}
\centering
\includegraphics[width=0.5\textwidth]{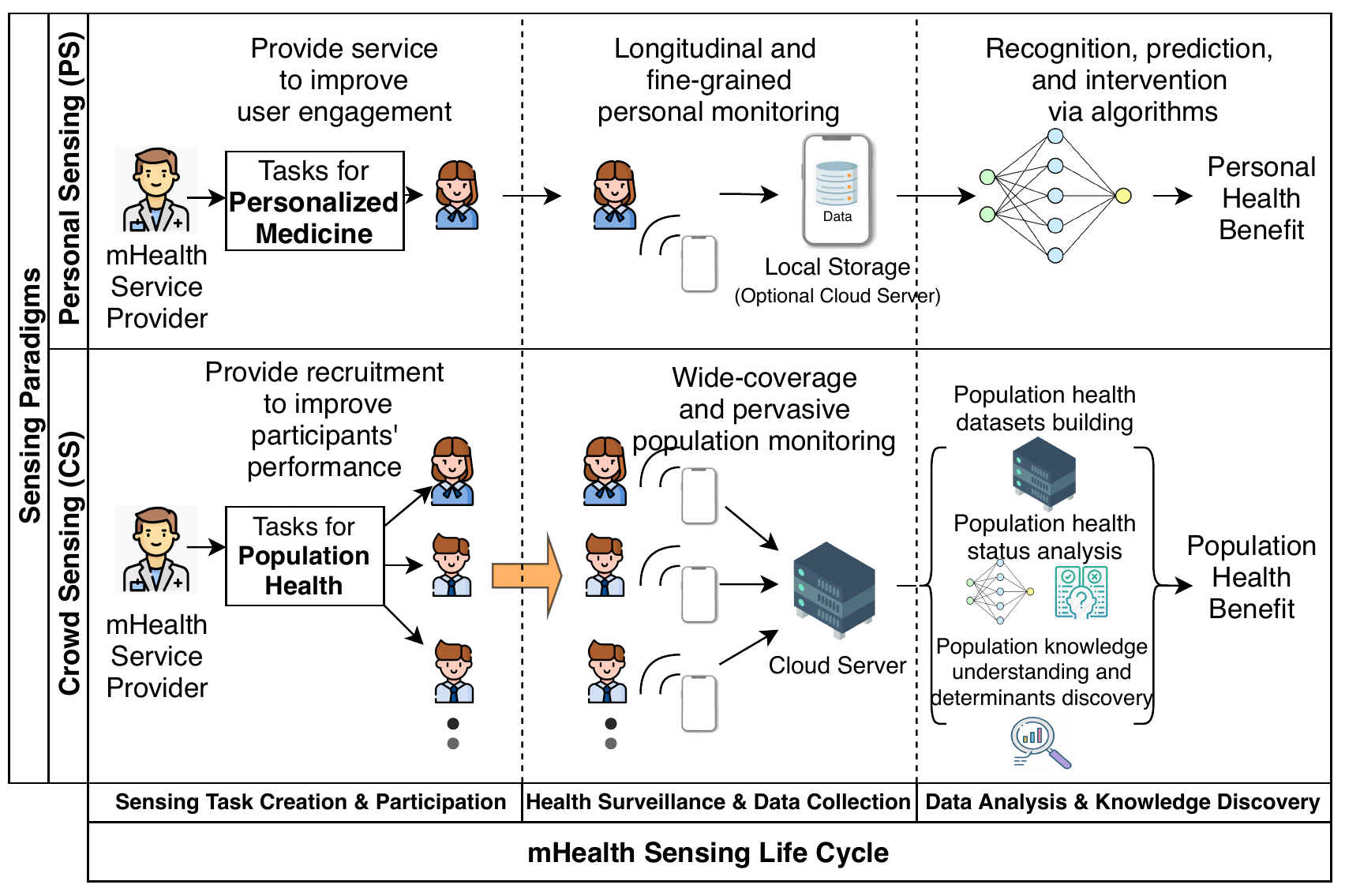}
\caption{The three-stage pipeline of mHealth Sensing apps}
\label{fig:taxonomy}
\end{figure}

\subsection{Objectives and Design \& Implementation (D\&I) Issues of mHealth Sensing Apps/Systems} 
\hl{For the two health outcomes of mHealth sensing, we plan to generalize and categorize existing works into two design paradigms -- \emph{\underline{(i) Personal Sensing} (PS)} and \emph{\underline{(ii) Crowd Sensing} (CS)} paradigms, according to the significant differences in app designs, such as user engagement strategies \cite{zhang2015incentives,wagner2017wrapper} and data analysis approaches \cite{huh2018big,liu2018large}. Moreover, as shown in Figure~\ref{fig:taxonomy}, we follow the common frameworks of mobile sensing apps~\cite{lane2010survey,zhang20144w1h} and propose to modularize the design of mHealth sensing apps~\cite{bardram2019the} for both Personal Sensing and Crowd Sensing into a pipeline of three stages as follows.}
\begin{enumerate}
\item \emph{Sensing Task Creation \& Participation - } With a pool of potential mobile users, the mHealth Sensing organizers create tasks for specific health issues via deployed apps~\cite{xiong2016sensus,hao2013isleep,abdullah2018sensing}, then prompt the participation of the users~\cite{lane2011bewell,wahle2016mobile,leao2016factors} or recruitment participants with incentives~\cite{xiong2017near,jaimes2017incentivization}.

\item \hl{\emph{Health Surveillance \& Data Collection -} With actively engaged participants, the mHealth Sensing apps and systems collect health-related data from participants in their daily life scenarios \mbox{\cite{rabbi2011passive,saeb2015mobile}}, then store and offload the sensing data with security and privacy-protection guarantees \mbox{\cite{wilkowska2012privacy,wang2013effsense,xiong2014emc,bertino2016data}}.}

\item \hl{\emph{Data Analysis \& Knowledge Discovery -} With health-related data collected, the mHealth Sensing apps and systems carry out data processing and analysis under ethical certification\mbox{\cite{albrecht2013transparency,giota2014mental,sharp2017mobile}} to predict health-related events for individuals\mbox{\cite{rabbi2015automated,agapito2016dietos}} and discover determinants of health{~\cite{kindig2003population}} -- i.e., knowledge about population health and well-beings\mbox{\cite{lane2011bewell,chen2021enabling}}.}
\end{enumerate}

\hl{Based on above two mHealth sensing design paradigms and the pipeline of three stages, this work provides two taxonomy systems that cover the major technical challenges and methodologies in this area. Specifically, we focus on the ``objectives'' (e.g., data privacy, data quality, energy efficiency, and other goals desired to enable practical mHealth sensing apps/systems) and ``designs \& implementations (D\&I)'' (e.g., methodologies to achieve the objectives) respectively. Furthermore, we review and categorize the sensing objectives and D\&I issues by the combination of two mHealth sensing design paradigms and three stages in details. Note that we discuss our motivations to structure these two taxonomy systems in Appendix.C.}

\subsection{Organization of the Survey}
\hl{The rest of this manuscript is organized as follows. Section \uppercase\expandafter{\romannumeral2} reviews several typical mHealth Sensing apps/systems for seven common health issues with case studies in details, where we specifically discuss ways mHealth sensing techniques handle the health issues for population health and personalized medicine purposes. Section \uppercase\expandafter{\romannumeral3} introduces the taxonomy system classifying mHealth sensing apps/systems from the \emph{sensing objectives} perspective. Section \uppercase\expandafter{\romannumeral4} presents the taxonomy system the classifies the mHealth sensing apps/systems from the perspective of \emph{sensing paradigms and D\&I issues}. In Section \uppercase\expandafter{\romannumeral5}, we point out identified research gaps and future directions in mHealth Sensing, while Section \uppercase\expandafter{\romannumeral6} concludes the article. In Appendix, we review the scientific approaches to this survey.}
\begin{table}[!htbp]
\caption{Terms and Definitions}
    \centering
    \begin{tabularx}{0.5\textwidth}{lX}
        \toprule
        \textbf{Terms}&  \textbf{Definitions}   \\
        \midrule
        Mobile Health
        & Namely \emph{mHealth}, a term used for the practice of medicine and public health supported by mobile devices \cite{adibi2015mobile}
        \\
        \midrule
        \hl{mHealth Provider}
        & \hl{Peoples (e.g., researchers, business companies) who organize healthcare services by designing and deploying mHealth techniques and applications \cite{leigh2019role}.}
        \\
        \midrule
        \hl{Ubiquitous Sensing}
        & \hl{Using networked sensors pervasively exist in daily-life scenarios to capture information about humans, environments, and their interactions~\cite{paulovich2018future}}
        \\
        \midrule
        Personal Sensing
        & A technique collecting and analyzing data from sensors embedded in the context of individual's life with the aim of identifying his/her behaviors, thoughts, feelings, and traits \cite{mohr2017personal}
        \\
        \midrule
        Crowd Sensing
        & A technique where a large group of individuals having mobile devices capable of sensing and computing collectively share data and extract information to measure, map, analyze, estimate or infer (predict) any processes of common interest \cite{ganti2011mobile}
        \\
        \midrule
        Passive Sensing
        & Sensing via devices that detect and respond to some type of input from the physical environment \cite{zhu2013sensec}
        \\ 
        \midrule
        Health Status
        & One's medical conditions (both physical and mental), claims experience, receipt of health care, medical history, genetic information, evidence of insurability, and disability \cite{belloc1972relationship}
        \\
        \midrule
        Health Outcomes
        & Health events occurring as a result of interventions \cite{manary2013patient}
        \\
        \midrule
        Health Determinants
        & Conditions which contribute to a wide range of health and quality of life-risks and outcomes \cite{marmot2005social}
        \\
        \midrule
        Mental Health
        & A state of well-being in which the individual realizes his/her own abilities, can cope with the normal stresses of life, can work productively and fruitfully, and is able to make a contribution to his/her community \cite{world2010mental}
        \\ 
        \midrule
        Treatment Effects
        & Causal effect of a given treatment or intervention on an outcome variable of health interests \cite{jacobson1999methods}
        \\
        \midrule
        Health Benefit
        & Positive phenomenons that a medicine treatment, substance or activity is improving health \cite{weinstein1996cost}
        \\
        \midrule
        Health Intervention
        & A treatment, procedure, or other action taken to prevent or treat disease, or improve health in other ways \cite{smith2015types}
        \\
        \midrule
        Digital Biomarkers
        & Objective, quantifiable physiological and behavioral data that are collected and measured by means of digital devices \cite{babrak2019traditional}
        \\
        \bottomrule
    \end{tabularx}%
    \label{definition}
\end{table}%

\section{Reviewing mHealth Sensing Apps and Systems for common health issues} \label{section:review}

In this section, we first list the definitions of some health-related terms in Table \ref{definition}. Then, with respect to seven most commonly researched health issues in reviewed papers, we review and summarize typical objectives and applications of mHealth Sensing works surrounding the seven issues (i.e., depression and anxiety, sleep quality and insomnia, diabetes, heart, elder-care, diet management, tinnitus, and COVID-19) from the two mHealth Sensing perspectives (i.e., \emph{(a) Personalized Medicine} and \emph{(b) Population Health}).

\begin{table*}[]
\caption{A summary table of the mHealth apps on seven common health issues created for the motivations in (a) Personalized Medicine and (b) Population Health respectively}
\centering
\begin{tabular}{|c|l|l|l|l|}
\hline
\multirow{2}{*}{\textbf{Health Issues}} & \multicolumn{2}{c|}{\textbf{Personalized Medicine}} & \multicolumn{2}{c|}{\textbf{Population Health}} \\ \cline{2-5} 
 & \multicolumn{1}{c|}{\textbf{Motivations}} & \multicolumn{1}{c|}{\textbf{Examples}} & \multicolumn{1}{c|}{\textbf{Motivations}} & \multicolumn{1}{c|}{\textbf{Examples}} \\ \hline
 
\multicolumn{1}{|c|}{\textbf{\begin{tabular}[c]{@{}c@{}}Depression\\and Anxiety\end{tabular}}} & \begin{tabular}[c]{@{}l@{}}Self-identifying and reducing\\ depression and anxiety\end{tabular} & \begin{tabular}[c]{@{}l@{}}\cite{mcintyre2021ecological}, \cite{burns2011harnessing},\\\cite{anthes2016mental}, \cite{hung2016predicting},\\\cite{asselbergs2016mobile}, \cite{wahle2016mobile}\end{tabular} & \begin{tabular}[c]{@{}l@{}}Population mental health screening\\ and determinants inferring \end{tabular} & \begin{tabular}[c]{@{}l@{}} \cite{kraft2020combining}, \cite{chow2017using},\\ \cite{saeb2015mobile}, \cite{boukhechba2017monitoring},\\ \cite{wang2016crosscheck}\end{tabular} \\ \hline

\multicolumn{1}{|c|}{\textbf{\begin{tabular}[c]{@{}c@{}}Sleep Quality \\ and Insomnia\end{tabular}}} & \begin{tabular}[c]{@{}l@{}}Monitoring and interventions\\ to promote sleep quality\end{tabular} & \begin{tabular}[c]{@{}l@{}}\cite{min2014toss}, \cite{hao2013isleep}, \\\cite{gu2014intelligent},\cite{sathyanarayana2016sleep}\end{tabular} & \begin{tabular}[c]{@{}l@{}}Population sleep statistics for \\understanding sleep science issues\end{tabular} & \begin{tabular}[c]{@{}l@{}}\cite{sharmila2020towards}, \cite{abdullah2014towards}\end{tabular} \\ \hline
\multicolumn{1}{|c|}{\textbf{Diabetes (type 2)}} & \begin{tabular}[c]{@{}l@{}}Glucose monitoring for type\\ 2 diabetes management\end{tabular} & \begin{tabular}[c]{@{}l@{}}\cite{sieverdes2013improving}, \cite{al2015mobile},\\ \cite{cappon2017wearable}, \cite{puhr2019real}\end{tabular} & \begin{tabular}[c]{@{}l@{}}Understanding the social determinants\\ contributing to diabetes\end{tabular} & \begin{tabular}[c]{@{}l@{}}\cite{boulos2021smart}, \cite{basatneh2018health} \end{tabular} \\ \hline

\multicolumn{1}{|c|}{\textbf{Heart}} & \begin{tabular}[c]{@{}l@{}}Heart rate monitoring and\\ heart disease prevention\end{tabular} & \begin{tabular}[c]{@{}l@{}}\cite{otoom2015effective}, \cite{triantafyllidis2015personalised},\\ \cite{ali2018real}, \cite{czosek2013utility},\\ \cite{koehler2018efficacy}, \cite{evans2016remote}\end{tabular} & \begin{tabular}[c]{@{}l@{}} Researching the impact of determinants\\ on cardiovascular diseases \end{tabular} & \cite{karhula2015telemonitoring} \\ \hline

\multirow{2}{*}{\textbf{Elder-care}} & \begin{tabular}[c]{@{}l@{}}In-home care service and\\ assistance\end{tabular}  & \begin{tabular}[c]{@{}l@{}}\cite{hsu2010rfid}, \cite{zhou2010case}\end{tabular} & \multirow{2}{*}{\begin{tabular}[c]{@{}l@{}}Understanding the health status and \\lifestyle of the elderly population\end{tabular}} & \multirow{2}{*}{\cite{marakkalage2018understanding}} \\ \cline{2-3}

 & \begin{tabular}[c]{@{}l@{}}Outdoor monitoring and\\ notification\end{tabular} & \begin{tabular}[c]{@{}l@{}}\cite{lin2012detecting}, \cite{du2008hycare}\end{tabular} &  &  \\ \hline
 
{\textbf{\begin{tabular}[c]{@{}c@{}}Diet and Weight\\ Management\end{tabular}}} & \begin{tabular}[c]{@{}l@{}}Diet self-monitoring and \\exercise management\end{tabular} & \begin{tabular}[c]{@{}l@{}}\cite{rabbi2015automated,agapito2016dietos}, \\ \cite{hezarjaribi2017speech2health,basatneh2018health}\end{tabular} & \begin{tabular}[c]{@{}l@{}}Understanding population eating \\patterns, episodes, and disorders\end{tabular} & \cite{meegahapola2020alone} \\ \hline

\textbf{Tinnitus} & \begin{tabular}[c]{@{}l@{}}Tinnitus self measurement \\and retraining therapy\end{tabular} & \begin{tabular}[c]{@{}l@{}}\cite{schlee2016measuring}, \cite{pryss2020applying}\end{tabular} & \begin{tabular}[c]{@{}l@{}} Studying symptoms, causes, and\\ treatments of tinnitus population \end{tabular} & \begin{tabular}[c]{@{}l@{}}\cite{pryss2015mobile}, \cite{mehdi2019towards},\\ \cite{pryss2017mobile}, \cite{probst2017does}\end{tabular} \\ \hline

\multirow{2}{*}{\textbf{COVID-19}} & \begin{tabular}[c]{@{}l@{}} Automatic self-diagnosis \end{tabular}& \begin{tabular}[c]{@{}l@{}}\cite{rao2020identification},\cite{sharma2020coswara},\\ \cite{brown2020exploring}, \cite{han2021exploring}\end{tabular} & \begin{tabular}[c]{@{}l@{}}Population screening the spread of\\ COVID-19\end{tabular} & \begin{tabular}[c]{@{}l@{}}\cite{xiong2020mobile,kang2020multiscale}, \\ \cite{grantz2020the,hao2020understanding}\end{tabular} \\ \cline{2-5} 
 & \begin{tabular}[c]{@{}l@{}}Contact tracing for infectious\\ risk estimation \end{tabular}& \begin{tabular}[c]{@{}l@{}}\cite{cho2020contact}, \cite{ienca2020on},\\ \cite{carli2020wetrace}, \cite{xia2020how}\end{tabular} & \begin{tabular}[c]{@{}l@{}}Public Health Policy Evaluation and \\Development \end{tabular}& \begin{tabular}[c]{@{}l@{}}\cite{domenico2020impact,vinceti2020lockdown,huang2020quantifying},\\ \cite{gao2020association,zhou2020effects,9378097},\\ \cite{gozzi2020estimating,Xiong2020.04.20.20068676,liu2021analysis}\end{tabular} \\ \hline
\end{tabular}
\label{app-summary}
\end{table*}

In general, as shown in Table \ref{app-summary}, for \emph{Personalized Medicine} objective, apps adopt \emph{Personal Sensing} paradigm and focus on individual's health benefit via D\&Is of health status monitoring, recognition, and intervention; while for \emph{Population Health}, the apps in a \emph{Crowd Sensing} manner mainly aim to measure/understand population health status and discover knowledge for public health benefit. 

Here we discuss three typical health issues among the seven from both perspectives of mHealth Sensing.

\subsection{Depression and Anxiety}

Depression and anxiety are the main mental health disorders, broadly experienced by 548 million people worldwide who hardly access effective treatment \cite{owidmentalhealth,bronson2017indicators}. mHealth Sensing gives ubiquitous and flexible solutions on both sides of personal mental health monitoring and intervention \cite{gravenhorst2015mobile,costa2017emotioncheck,boonstra2018using,rey2018personalized,abdullah2018sensing,tseng2021digital} and population mental health surveying and understanding \cite{torous2015realizing,wang2020social}.

\paragraph{Personal Depression and Anxiety Monitoring and Intervention}

mHealth Sensing techniques are providing broadly accessible services for individuals with mental disorders, as they can collect user's daily-life indicators varying with time and scenarios, as well as deliver timely interventions though without scarce clinical resources \cite{boukhechba2020leveraging}. A typical application of Personal Sensing for depression and anxiety is Mobilyze! \cite{burns2011harnessing}, a mobile mental intervention application with a two-step framework -- context sensing and ecological momentary intervention. By collecting contextual data such as locations, recent calls, ambient light, and feeding them into a medical diagnosis model, it infers user's mental health status and provides interventions to guide to overcome psychological dilemmas (e.g., lack of social interaction). Furthermore, in mental health domain, the concept of just-in-time adaptive intervention (JITAI) was proposed to guide timely and personalized interventions \cite{nahum2018just}. For example, advances in artificial intelligence are promoting smarter decision-making of when and where it is most helpful to provide supportive interventions by learning from individual's historical behaviors \cite{menictas2019artificial,menictas2020fast}.

\paragraph{Population Depression and Anxiety Survey}
Mobile Sensing techniques are increasingly being adopted to population depression and anxiety surveys, as they provide a low-cost (both in resources and time), widespread, and online data collection manner versus laborious and high-cost clinical testing and questionnaires. For example, by studying the correlation between anxiety and behavioral indicators (e.g., activity locations, text messages, and calls) in a 54-students group over two weeks, Boukhechba \emph{et al.} \cite{boukhechba2017monitoring} proposed flexible anxiety assessment methods for monitoring college students via mobile apps.

\paragraph{Population Mental Health Determinants Understanding} 
New inspirations and knowledge about population mental health determinants can be gained via massively collecting and comparatively analyzing data among populations \cite{kraft2020combining}, such as inferring causes between social anxiety and group behavioral patterns \cite{vanderweele2016causal}. For example, a Mobile Crowd Sensing platform -- Sensus \cite{xiong2016sensus} was leveraged by Chow \emph{et al.} to verify clinical models of depression and anxiety \cite{chow2017using}. \hl{Taking the levels of depression and social anxiety as moderators, researchers tested the relations between state effect and time spent at home of 72 recruited students, and finally, they gain an understanding on the significant correlations between depression \& anxiety and home-stay behaviors in the target population.}

\subsection{Sleep Quality and Insomnia}

mHealth sensing applications are widely applied to monitor sleep status and measure sleep quality. The basis for this is that the digital biomarkers (e.g., heart rate and sound of snoring) related to sleep can be easily collected by mobile sensors during sleeping \cite{surantha2016internet}.

\paragraph{Personalized Sleep Monitoring and Insomnia Assistance}
mHealth Sleeping apps are giving more and more accessible sleep quality monitoring and sleep-aid services to users \cite{chen2013unobtrusive,min2014toss,surantha2016internet,sathyanarayana2016sleep}. Several sleep monitoring systems are deployed on wearable devices requiring users to wear a product embedded with specific sensors during sleeping, which is either limited in clinical environment \cite{de2019wearable} or uncomfortable to the users \cite{looney2016wearable}. A new trend in mHealth Sensing for sleep monitoring is using off-the-shelf mobile phones built-in sensors such as microphones and accelerometers to detect the sleep duration and infer sleep quality. For example, Hao \emph{et al.} \cite{hao2013isleep} proposed to leverage microphone audio to detect the events closely related to sleep quality such as ambient noise, body movement, and snoring \cite{min2014toss} to enable personalized and in-place sleep quality monitoring. Furthermore, Gu \emph{et al.} \cite{gu2014intelligent} mined and detected the sleep stage (e.g., week sleep, deep sleep, and rapid eye movement) by monitoring sleep environment and personal factors leveraging a statistical model, which provides fine-grained descriptions of sleep status.

\paragraph{Population Sleep Science Research}
Crowd Sensing apps are widely used to create population sleep status datasets for sleep science, such as understanding the issues on psychological research and sleep science, as one's sleep quality interacts with her/his lifestyle and mental status. In practice, to understand the behavioral pattern between phone usage and sleep quality, recently, Sharmila \emph{et al.} \cite{sharmila2020towards} collected a large-scale phone usage dataset and sleep questionnaires from 743 participants of different ages and socioeconomic backgrounds in a Crowd Sensing manner and figure out the effect of mobile phone usage patterns on sleep using statistical methods. Abdullah \emph{et al.} \cite{abdullah2014towards} proposed to study the effects of sleep quality on people’s daily rhythm and well-being including levels of alertness, productivity, physical activity, and even sensitivity to pain.

\subsection{Mobile Sensing in COVID-19 Era}

The mobile devices that people carry around are like "witnesses" to the spread of the epidemic, as the spread of the COVID-19 virus is accompanied by human mobility and contact, where Mobile Sensing have shown its great power in COVID-19 era \cite{pan2020application}; the typical contributions include personal diagnosis \cite{rao2020identification}, infection traceability \cite{cho2020contact}, transmission interpretation \cite{grantz2020the} and policy decision-making \cite{oliver2020mobile}, etc.

\paragraph{Personal Automatic Self-Diagnosis via Sounds}
Mobile microphones collect audio samples such as sighs, breathing, heart, digestion, vibration sounds on body, which can serve as the indicators to diagnose lung diseases \cite{chamberlain2016application}, giving great possibilities of automatic detection and diagnosis of COVID-19 infection \cite{sharma2020coswara}. For example, Brown \emph{et al.} \cite{brown2020exploring} proposed methodologies to detect diagnostic signs of COVID-19 from voice and coughs, which well distinguish a user who is COVID-19 positive with a cough from a negative user with a cough. In addition to voice analysis, Han \emph{et al.} \cite{han2021exploring} further explored fusion strategies to combine voice and reported symptoms which yield better detection performance.

\paragraph{Personal Contact Tracing}
The COVID-19 virus spreads from an infected person's mouth or nose in small liquid particles when they cough, sneeze, speak, sing or breathe \cite{world2020coronavirus}, causing finding the contacts of positive patients is an essential task for epidemic control. Many contact tracing mobile apps are developed and deployed for privacy-preserving and comprehensive COVID-19 tracing for individual users to check their contact history with mobile phone data \cite{cho2020contact,ienca2020on,xia2020how,xu2021beeptrace}. For example, Carli \emph{et al.} \cite{carli2020wetrace} developed WeTrace, a mobile COVID-19 tracing app which detects and records one's contact with others leveraging the interaction via Bluetooth Low Energy (BTE) communication channel; and a trusted data transmission framework is proposed to balance the health and the privacy perspectives.

\paragraph{Epidemic Spreading Analysis via Human Mobility}
The significant correlation between human mobility and COVID-19 infections provides guidance on investigating and understanding the spreading of COVID-19 via multi-scale human mobility data \cite{xiong2020mobile,kang2020multiscale,grantz2020the}. From the perspective of the human mobility research, large-scale and long-term GPS data can be used to detect high-risk regions \cite{dephillipo2021mobile}, and population traveling data (e.g., Baidu Qianxi \cite{baidu}) can be leveraged to analyze the spreading path between cities and countries \cite{arun2020bluble}. For example, by incorporating human mobility data into epidemic modeling, Hao \emph{et al.} \cite{hao2020understanding} studied how the multi-scale urban human mobility impacts the spreading process at varying levels, which provides insights on making smarter policies to respond the next outbreak.

\paragraph{Public Policy Evaluation and Making}
Strict infection control policies proposed by governments have been taken to limit and mitigate the fast-spreading of COVID-19, such as lock-down, travel restrictions, quarantine, social distance ban. mHealth Sensing data among populations is contributing in evaluating and making these policies \cite{kishore2020measuring,oliver2020mobile}. Intuitively, several of the sensing indicators among populations, such as the average time of users stay-at-home and the number of mobile devices in a public place, can be used to measure stay-at-home and social distance policy efficiency \cite{domenico2020impact,vinceti2020lockdown}. Furthermore, statistical and machine learning methods can be used to estimate, simulate, and predict the effects of the policies on controlling virus spreading driven by the population data gathered in mobile devices \cite{gao2020association,zhou2020effects,gozzi2020estimating}.

\subsection{Discussion}
Note that, in this work, we review and summarize the works on mHealth Sensing Apps and Systems that deployed over massive smartphones and commodity interactive devices, such as tablets, smartwatches, and other wearable consumer electronics in non-invasive sensing manners. Many other works intending to monitor physiological status of patients for medical purposes or professional devices/systems for critical cares/assisted living, such as medical sensors~\cite{beck2000remote,chen2014continuous,khakh2006p2x,guntner2019breath,gao2020flexible}, Internet of Medical Things (Medical IoTs) and Medical Cyber-Physical Systems (Medical CPSs)~\cite{gatouillat2018internet,dimitrov2016medical,haghi2017wearable,hayani2020image,joyia2017internet,elhoseny2019effective,pazienza2020adaptive,limaye2018hermit,gayathri2019efficient,atat2018a,haghi2020a,rahman2018m,jiang2019toward,parah2020efficient,peng2020secure,liu2020improved,zhang2019energy,meng2021hybrid,amendola2014rfid,lin2020covid,sundaravadivel2018everything,hossain2018an,zhu2015bridging
}, and medical robots~\cite{altaee2017robot,taylor2003medical,webster2006nonholonomic,wang2011lower,beasley2012medical,fukushima2014medical}, are not included here. \hl{Of-course, there are many other behavior-related health issues that are not well covered here, such as drug/alcohol abuses or addiction~\cite{phan2019drinks,Tamersoy:2015:CSD:2700171.2791247} in general.}

\begin{figure*}
\centering
\includegraphics[width=0.8\textwidth]{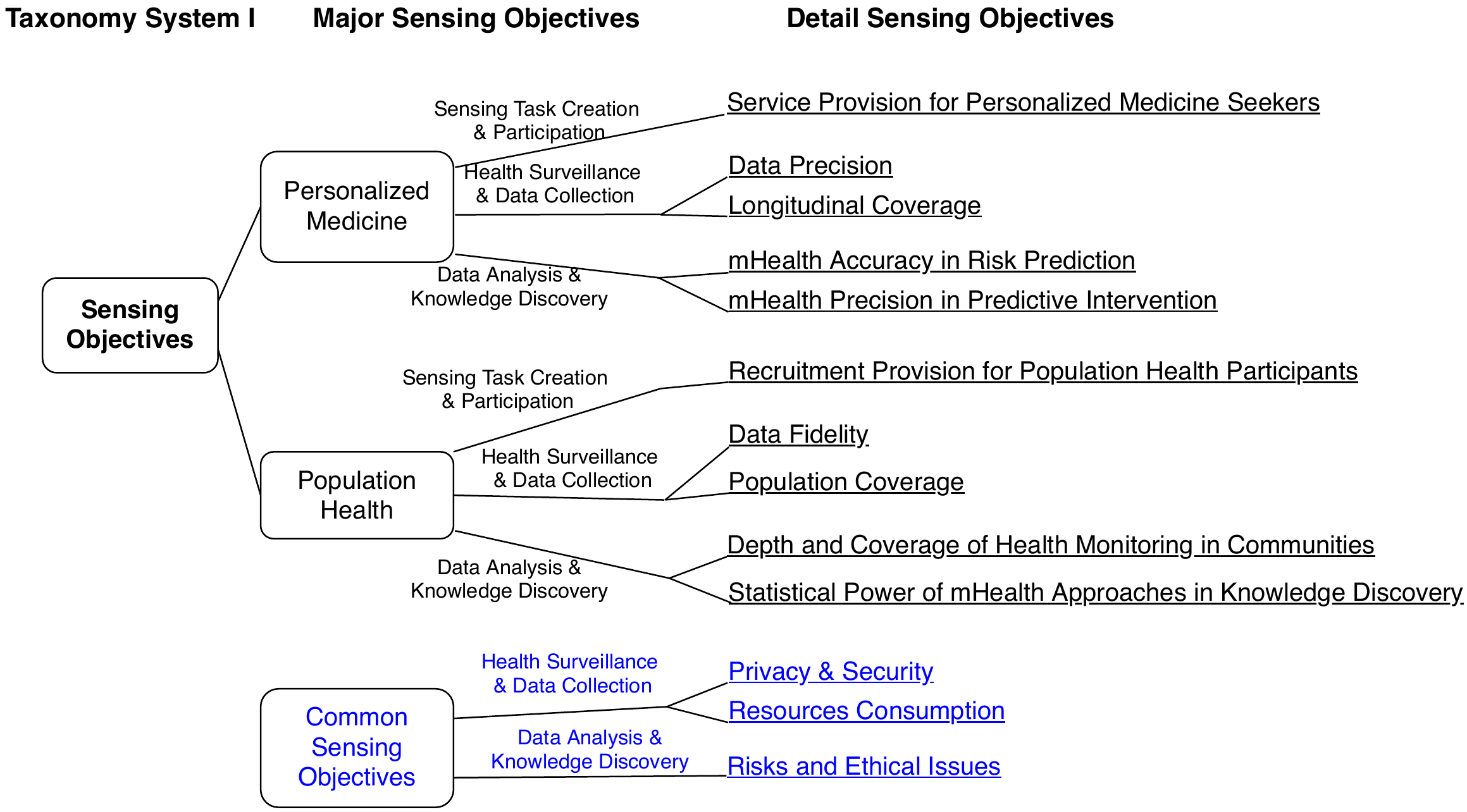}
\caption{Taxonomy System \uppercase\expandafter{\romannumeral1} -- Sensing Objectives}
\label{fig:taxonomy-system-1}
\end{figure*}

\section{Taxonomy System \expandafter{\romannumeral 1}: Classification of mHealth Sensing by Sensing Objectives} \label{section:objectives}

\hl{In this phrase, we introduce the proposed mHealth Sensing taxonomy system { \uppercase\expandafter{\romannumeral1}} from \emph{sensing} \emph{objectives} perspective. With respect to the two perspectives of modern healthcare, as shown in Figure{~\ref{fig:taxonomy-system-1}}, we specify and classify \emph{Major} \emph{Sensing} \emph{Objectives} of mHealth Sensing apps as \emph{(a)} \emph{Personalized} \emph{Medicine} and \emph{(b)} \emph{Population} \emph{Health} apps. Then we further discuss the \emph{Detail} \emph{Sensing} \emph{Objectives} in each step of the life-cycle framework of mHealth Sensing.}

\subsection{Objectives in Sensing Task Creation \& Participation} \label{section:ob-participation}
The main objectives in Sensing Task Creation \& Participation step are creating and allocating health-related tasks in mobile apps, then prompt the participation of the users or recruited participants to execute the sensing tasks. Since the health benefit for the participants in the two types of apps varies (i.e., participants in \emph{Personalized Medicine} apps obtain direct personal health benefit, while participants in \emph{Population Health} practices hardly obtain health benefit equaling to their efforts), where the detail objectives in this step can be distinguished as \emph{service provision for personalized medicine seekers} and \emph{recruitment for population health participants}.

\begin{itemize}
\item \textbf{Service Provision for Personalized Medicine Seekers -} Sensing apps for \emph{Personalized Medicine} provide accurate health status monitoring and personalized interventions or treatments, which can be concluded as healthcare \emph{services} provision \cite{riley2011health,nahum2018just,anjomshoaa2018city}. In most of the \emph{Personalized Medicine} cases, participants actively engage in the sensing task for personalized medicine with an expectation to seek and extend personal health benefit \cite{klasnja2009using}. To this end, the detail objective of personalized medicine apps in this step is to provide exact healthcare services (e.g., exercise reminders and user-friendly interface) and keep improving service quality (e.g., optimizing intervention times with algorithms) to guarantee and enhance users' active engagement \cite{asimakopoulos2017motivation}.

\item \textbf{Recruitment Provision for Population Health Participants -} \hl{\emph{Population Health} apps are mostly for studying population health issues leveraging massive collected data from groups, causing a problem for participants is that -- there is no intuitive and sufficient health benefit gained for themselves to compensate for their costs and concerns (e.g., time consumption, privacy exposure{~\cite{spathis2019sequence}}, and battery usage{~\cite{boukhechba2020swear}})}. For example, in a COVID-19 infectious population screening{~\cite{rashid2020covidsens}} or a rare clinical disease causes understanding program{~\cite{ben2020mobile}}, the results are valuable for organizers but limited for participants. The above reasons lead to a unique detail objective of Population Health apps in absorbing participation -- providing recruitment to gather participants and motivating their performance with incentives \cite{peng2016qualitative,jaimes2017incentivization}.
\end{itemize}

\subsection{Objectives in Health Surveillance \& Data Collection}

\begin{figure}
\centering
\includegraphics[width=0.5\textwidth]{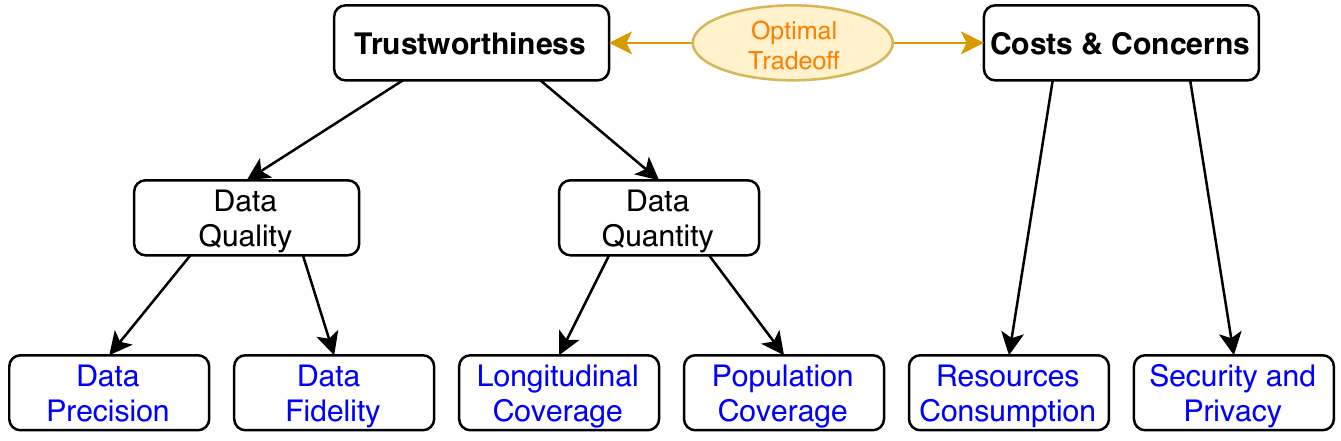}
\caption{Detail sensing objectives in Health Surveillance \& Data Collection}
\label{fig:data-collection-objectives}
\end{figure}

With exact sensing tasks and a pool of users/participants, the bottleneck in \emph{Health Surveillance \& Data Collection} is -- how to effectively collect and gather trustworthy sensing data, with taking users' costs and concerns into consideration. As Figure \ref{fig:data-collection-objectives}, we summarize that mHealth Sensing apps' trustworthiness lies in \emph{data quality} and \emph{data quantity}; further, the \emph{data quality} can be further 
indicated as \emph{data precision} and \emph{data fidelity}, and the \emph{data quantity} can be divided into \emph{longitudinal coverage} and \emph{population coverage}. 
\hl{In addition, some objectives are commonly existed in both Personalized Medicine and Population Health practices, such as \emph{security \& privacy} and \emph{resources consumption}, but vary in details, where we reviews these issues in the last of this section.}


\paragraph{Personal Sensing for Personalized Medicine} 
In \emph{Personalized Medicine} mobile apps, to provide timely and adaptive healthcare services based on precise and sufficient data, the detail objectives in this step are \emph{data precision} and \emph{longitudinal coverage} in data collection process.

\begin{itemize}
    \item \textbf{Data Precision -} The data precision is the most straightforward pursuit of Personal Sensing tasks, which determines the service quality of Personalized Medicine. Here we give the mobile medical devices in the intensive care unit (ICU), which is the last barrier to save the lives of dying patients in the hospital, as examples \cite{halpern2016trends,ahouandjinou2016smart}. The personal wearable devices with incentive body sensors, light and sound sensors, and others precisely collecting the physical and environmental context data from ICU are typical Personal Sensing schemes with high sampling precision, finely sensing physical and environmental biomarkers such as facial expressions, functional status entailing extremity movements and postures, and environmental indicators for ICU’s context \cite{davoudi2019intelligent}.
    
    \item \textbf{Longitudinal Coverage -} Data with large longitudinal helps understanding personal health issues comprehensively for two fold reasons -- not only longitudinal moment-to-moment data sampling is helpful for capturing complex health dynamics to achieve meaningful modelling and prediction \cite{roshanaei2009longitudinal,servia2017mobile,fong2017longitudinal,kalanadhabhatta2021effect}, but also the analyzing of the onset of some diseases is not trivial, as the disease might be triggered by the interaction of multiple pathogenic determinants over a long period (e.g., monthly and annually), which cannot be detected with a brief observation \cite{harari2016using,mcnamara2016sadhealth,majumder2019smartphone}. \hl{In addition, the interaction of mHealth Sensing apps is also beneficial to the enlargement of longitudinal coverage of data collection, which shares and gathers information between apps. For instance, Google Health and HealthVault are cross-platform personal health record systems storing and sharing information between mHealth apps in a secure and privacy-protected manner, which gives mHealth apps a great potential for comprehensively serving health and well-being{~\cite{sunyaev2010evaluation}}.}

\end{itemize}

\paragraph{Crowd Sensing for Population Health} In \emph{Population Health} practices, the task of Health Surveillance \& Data Collection is to build a large-scale and error-free data pool surrounding the health issues to be analyzed and researched, with detail objectives of ensuring \emph{data fidelity} and enlarging \emph{population coverage} in the sensing process.

\begin{itemize}
    \item \textbf{Data Fidelity -} Versus data precision, data fidelity in the mHealth Sensing context refers to that there is no human error (e.g., intentional cheating or equipment failure) in the gathered data \cite{gilbert2010toward,rahman2017mdebugger,rahman2019towards}. Especially, different from the collections of some general datasets (e.g., traffic speed data or urban temperature data) which can be gathered in a short time, collecting daily/clinical health-related data requires enormous manpower, incentive cost, and devices resources in a long time \cite{feldman2012big,khozin2017real}. Also, once human errors are introduced into the data pool, it would causes biased health modeling, inaccurate treatment effect measurement, and wrong medical conclusions, which are harmful to the health and well-being purposes \cite{bell2020frequency}.
    
    \item \textbf{Population Coverage -} Enlarging the population coverage of Health Surveillance \& Data Collection is beneficial to obtain statistical-significant and generalized Data Analysis \& Knowledge Discovery. Specifically, in Crowd Sensing paradigm, some general guarantees for population coverage are age, gender, region, patient groups coverage; for varied research purposes, the coverage requirements for population attributes vary \cite{trifan2019passive,zhang2021passive}. For example, data for population mental health researches should cover balanced genders and diversified ages for comparative analysis and knowledge discovery with no/limited prior knowledge leveraging machine learning \cite{wiens2018machine} or statistical inference \cite{edwards1963bayesian} approaches; data for sleep science researches should cover kinds of patient groups such as sleep apnea, insomnia, Parkinson's disease, and \hl{periodic limb movement disorder (PLMD)}, as well as healthy people as control group.

\end{itemize}

In addition, though the detail sensing objectives in Health Surveillance \& Data Collection step are specified as the above perspectives, these objectives are usually overlapped. For example, \emph{data precision} and \emph{longitudinal coverage} are also meaningful in \emph{Crowd Sensing for Population Health} practices, but compared to these two objectives, \emph{data fidelity} and \emph{population coverage} are in need of relatively dedicated D\&Is for specific existing problems.

\paragraph{Commonly Existed Objectives -- Concerns \& Costs}

\hl{Beyond the technical objectives in trustworthiness of data, other issues in solving users' practical concerns and costs are the common objectives for mHealth Sensing apps.}

\begin{itemize}

    \item \hl{\mbox{\textbf{Security \& Privacy -}} Issues in security and privacy are greatly concerned in health-related domains, as health data is top sensitive{~\cite{wilkowska2012privacy,xu2021beeptrace}}. As for Personalized Medicine mobile apps, the security/privacy issues include identity privacy{~\cite{he2015user}} (participants do not want to expose personal information), data privacy{~\cite{wu2017dynamic}} (health-related data is top sensitive), attribute privacy{~\cite{ni2017privacy}} (for attributes such as locations and trajectories). Besides, the risk of privacy leakage in Population Health apps is greater~\cite{martinez2015privacy,papageorgiou2018security}, as it requires regular sensitive health-related data uploading and offloading between mobile devices and cloud servers via networks{~\cite{wang2020privacy}}. To be specific, additional privacy concerns in Population Health data collecting and uploading processes are task privacy~\cite{pournajaf2014survey} (the sensing tasks may correlate to participants' illnesses), and decentralized privacy~\cite{ma2019decentralized} (frequent communication with a central server could be more easily hacked).}

     \item \hl{\mbox{\textbf{Resources Consumption -}} Keeping mobile sensing data sampling causes considerable battery, hardware, and software resources consumption, for every mHealth Sensing participants. From \mbox{\emph{Personalized Medicine}} perspective, the resource consumption is more intense, as its data collection actions are generally continuous and intensive \mbox{\cite{balan2014challenge}}. Against this background, the type and combination of sensors in working and their sampling rate, data accuracy and sampling abundance are under consideration \mbox{\cite{ahnn2013mhealthmon,wu2017reducing}}. 
     From \mbox{\emph{Population Health}} perspective, when the hardware consumption of each individual's perception is already relatively economical, the decrease in resources consumption are mainly achieved optimizing the task allocation in spatial, temporal, participants, and content to achieve cost-effective globally sensing \mbox{\cite{xiong2015eemc,wang2016fine,wang2018multi,wang2018energy,chen2021enabling}}.
}
\end{itemize}

\hl{Worth mentioning, the pursuit of data \emph{trustworthiness} may increase the \emph{concerns \& costs} of users; also, \emph{concerns \& costs} also limit the intensive, longitude, and broad-coverage data sampling of users. It leaves app developers to make an \emph{optimal tradeoff} between the two objectives in practice, as shown in Figure{~\ref{fig:data-collection-objectives}}. On the one hand, the developers should design and develop the apps with certifications that minimize data access privileges subject to the actual needs, to release the \emph{costs \& concerns}. On the other hand, advances in resources saving and privacy protection approaches may make it possible for developers to obtain additional permissions from users, which further improves the apps' \emph{trustworthiness}.}


\subsection{Objectives in Data Analysis \& Knowledge Discovery} \label{section:ob-data-analysis}

After gathering expected personal or population data pools, the main objective in Data Analysis \& Knowledge Discovery fold is to discover health-related knowledge about individuals and populations from gathered data, and provide adaptive and timely healthcare as feedback \cite{jothi2015data,olaronke2016big}. 

\begin{figure}
\centering
\includegraphics[width=0.5\textwidth]{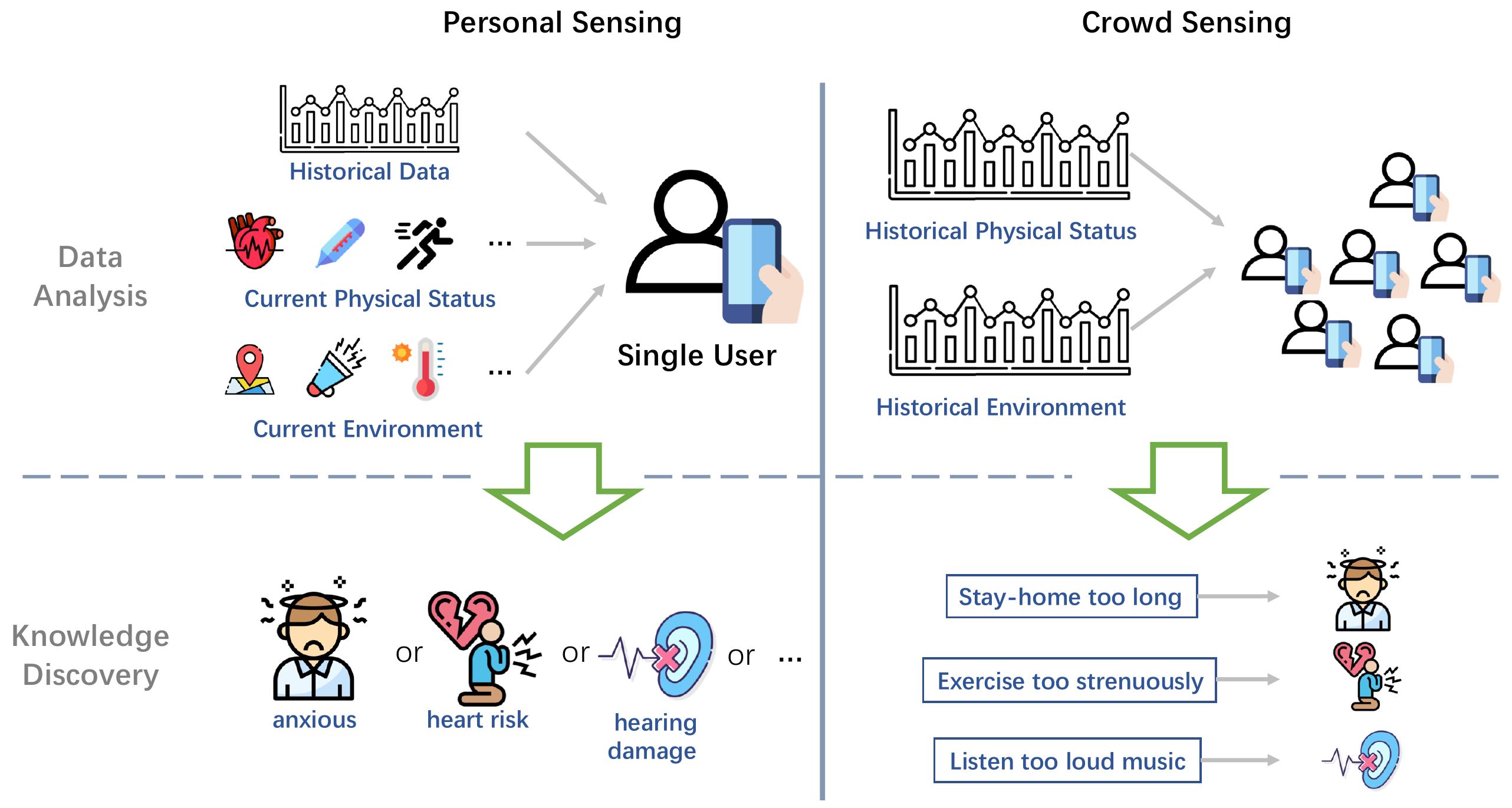}
\caption{The comparison of main objectives of Data Analysis \& Knowledge Discovery in the apps for Personalized Medicine and Population Health}
\label{fig:purpose-data-analysis}
\end{figure}

\paragraph{Personal Sensing for Personalized Medicine} Apps for Personalized Medicine usually recognize \cite{wood2019taking} or predict \cite{boukhechba2018predicting,tseng2020using} individual user's health status by integrating his/her historical, as well as current physical and environmental data surrounding a specific health issue to accurately recognize/predict health risks and provide precise healthcare interventions at the right time, as shown in Figure \ref{fig:purpose-data-analysis}.

\begin{itemize}

    \item \textbf{mHealth Accuracy in Risk Prediction -} Effective personalized healthcare services rely on the accuracy in the health status modeling and progression prediction. Sufficient multimodal data collected user's daily life such as self-reported medical history, physical biomarkers (e.g., heart rate), and environmental biomarkers (e.g., locations) provides great information for accurately modeling and predicting one's health outcomes and progressions via machine learning approaches \cite{barrett2020mobile,sedgley2013responsibilities,fragala2019population}. For example, by passively monitoring schizophrenia patients’ psychiatric symptoms represented by 7-item scale scores and behavioral/contextual characteristics (e.g., physical activity, conversation, mobility) over months, Wang \emph{et al.} \cite{wang2017predicting} proposed a prediction system which predicts psychiatric symptoms' dynamics and progression merely based on mHealth Sensing data without traditional self-reported ecological momentary assessment (EMA).
    
    \item \textbf{mHealth Precision in Predictive Intervention -} A typical detail objective in this step for Personalized Medicine apps is to provide predictive interventions with high mHealth precision responding to recognized/predicted health outcomes and progressions (e.g., increasing depression and anxiety, exposing to high heart risk, and being damaged hearing).
    Specifically, the precision above lies on precise intervention timing, measures, and intensity, which leads to just-in-time, adaptive, and effective mHealth supporting services \cite{nahum2018just,bidargaddi2020designing}. For example, Costa \emph{et al.} \cite{costa2017emotioncheck} proposed to improve one's cognitive performance by unobtrusively regulating emotions with smartwatch notifications in varying detected heart rates. Lei \emph{et al.} \cite{lei2017actor}, by formulating the intervention tasks in real-time as a contextual bandit problem, provided an online actor-critic algorithm as an intervention strategy to guide JITAI practices.
\end{itemize}

\paragraph{Crowd Sensing for Population Health} Crowd Sensing practices investigate population health issues by comprehensively mining massive health-related data among researched groups such as monitoring and screening the population health status in a region in both depth and coverage \cite{fragala2019population,jovanovic2019mobile}, and verifying \cite{sharmila2020towards} and inferring \cite{wang2016crosscheck} the determinants of specific diseases via powerful statistics-based approaches.

\begin{itemize}
    \item \textbf{Depth and Coverage of Population Health Monitoring in Communities -} For Population Health apps (especially for the apps on population health monitoring, screening, and surveying), in terms of data analysis, it is meaningful to deeply mine and widely enlarge the information of targeting communities leveraging collected Crowd Sensing data. For example, in many mHealth Crowd Sensing practices, some specific characteristics of health problems (e.g., the contact infection of infectious diseases \cite{clayton1993spatial}, familial heredity phenomenon of genetic diseases \cite{seaquist1989familial}, and regional relevance of conventional health habits \cite{cecilia2020mobile}) give great possibility to finish a mobile population health screening of the whole community by only investigating a subset of this group, which is a manner with accuracy guarantee and lower cost.
    
    \item \textbf{Statistical Power of mHealth Approaches in Knowledge Discovery -} The statistical power of the mHealth approaches is a key pursuit for medical-related knowledge discovery in large-scale population data. Specifically, in mHealth field, Crowd Sensing is being used as a useful tool to collect and analyze massive population health-related data to obtain medical knowledge, where new knowledge can be summarized or inferred by statistical methods for a better understanding of health determinants \cite{marmot2005social}, such as staying home too long causes mental health problems \cite{boukhechba2017monitoring}, lacking exercise would increase the risk of heart attack \cite{ali2018real}, and listening too loud music leads to tinnitus \cite{pryss2020applying}. For example, Zhang \emph{et al.} \cite{zhang2021passive} revealed how human mobility features extracted from large-scale human mobility data affect one's health conditions and which group of features contribute significantly leveraging statistical approach -- shapely additive explanation value analysis, which shed light on how to understand human mobility data in health monitoring domain.
\end{itemize}
\paragraph{Commonly Existed Objectives -- Risks and Ethical Issues}

\hl{In both mHealth-based \emph{Personalized Medicine} and \emph{Population Health} knowledge discovery practices, some risks and ethical issues cannot be ignored in sensing objectives.}

\begin{figure}
\centering
\includegraphics[width=0.48\textwidth]{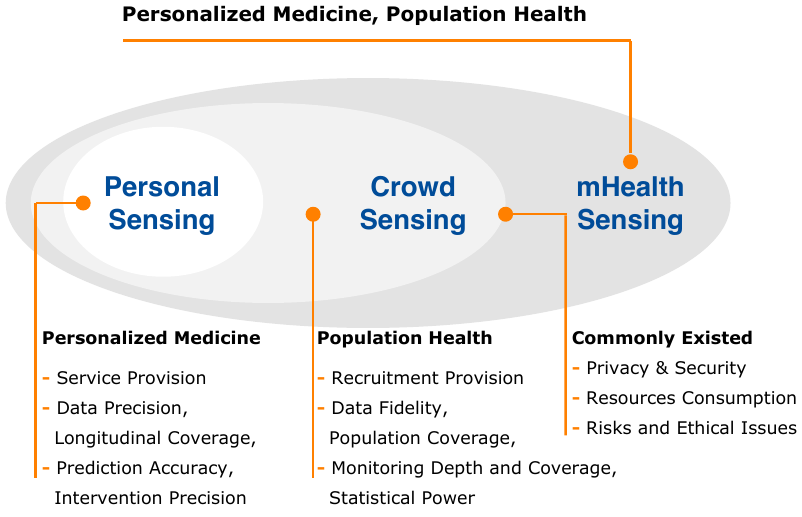}
\caption{The relationship between the objectives of Personal Sensing and Crowd Sensing paradigms}
\label{fig:objective-ps-cs}
\end{figure}

\begin{itemize}

    \item \hl{\mbox{\textbf{Risks and Ethical Issues -}} Risks and ethical issues are crucial in human-subject and mHealth research, since personally identifiable health-related data of users would be collected, uploaded, and analyzed, as well as sensitive scientific study results would be made public to varying degrees, even if some certifications are issued by the developers \mbox{\cite{rothstein2015ethical,yip2016legal}}. For instance, funded by three-party advertisers, such as insurance companies, developer may exposure information to them; some patients and victims may be forced to pay more or even fail to apply, which goes against ethics{~\cite{liu2015data}}. Besides of revealing private health information, common risks and ethical issues in mHealth Sensing apps include data loss, theft and hacked{~\cite{luxton2011mhealth}}, excessive or unauthorized collection of data{~\cite{kang2020systematic}}, loose medical conclusions and negative impact on life \mbox{\cite{wakefield2007concept,scott2015review}}. Besides, the scientific studies carried with mHealth apps may be not solid enough, since most of the obtained conclusions are based on limited observation samples and periods; for example, few studies have conducted follow-up studies on large-scale populations for more than a few months, and exact long-term impact of mHealth sensing apps on personal and population health is still not scientifically clarified{~\cite{fiordelli2013mapping}}. Against this background, appropriate analysis of potential risks{~\cite{scott2015review}}, ethical issues\mbox{\cite{mittelstadt2017designing,mittelstadt2017ethics}}, as well as previously mentioned security \& privacy issues should be done ahead of issuing certifications of mHealth apps being used in daily-life and even medical scenarios.}
    
\end{itemize}

It is worth mentioning that, applying Crowd Sensing apps can be regarded as a accumulation of the number of Personal Sensing apps deployed in a community. Thus, most of the objectives in Personal Sensing are also what the Crowd Sensing paradigm pursues in practice. To this end, here we conclude the objectives of Personal Sensing, Crowd Sensing, and mHealth Sensing as shown in Figure \ref{fig:objective-ps-cs}, where their objectives are progressive. For example, intuitively, in Crowd Sensing practices, improvements in cost saving and data accuracy also certainly prompt the performance of the apps.

\begin{figure*}
\centering
\includegraphics[width=0.8\textwidth]{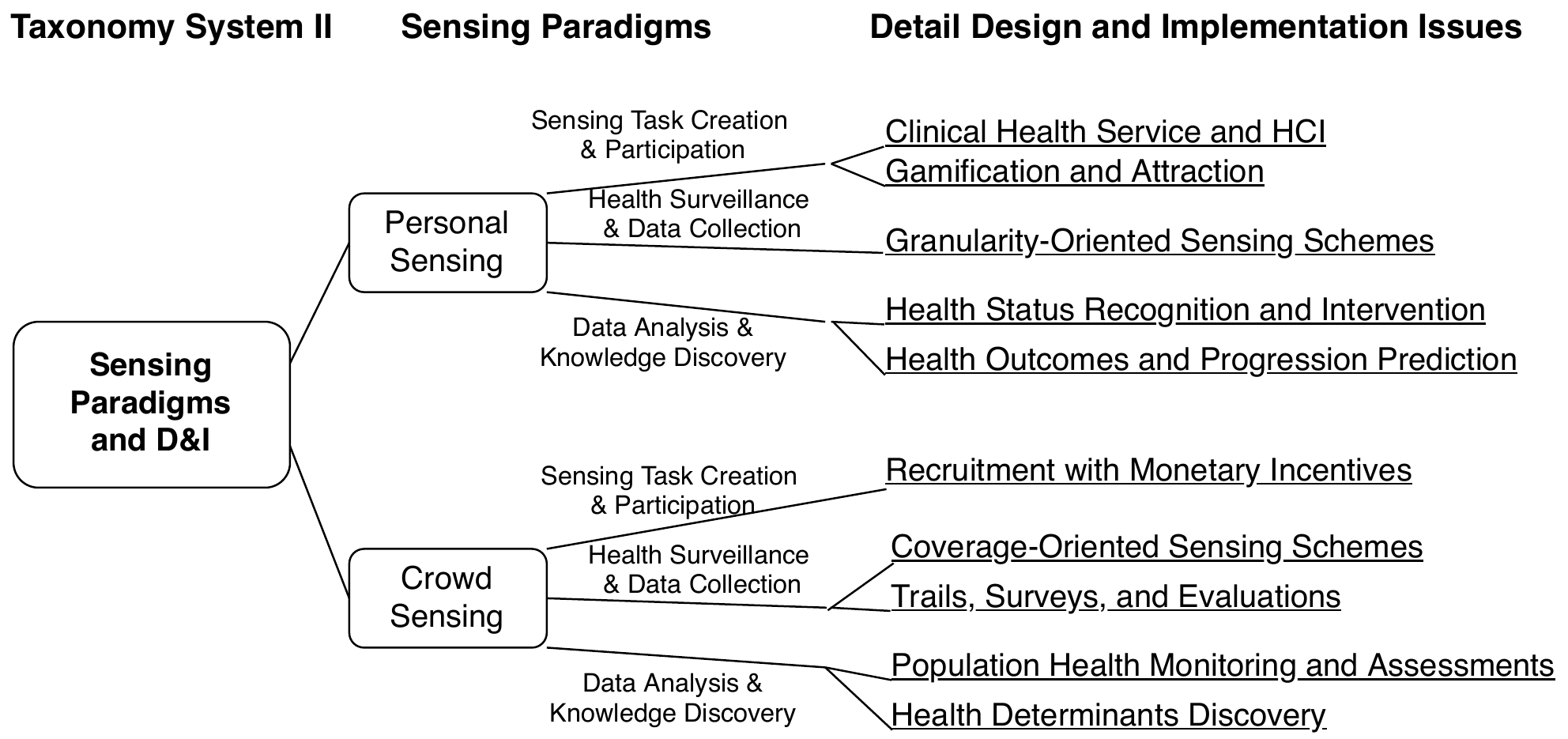}
\caption{Taxonomy System \uppercase\expandafter{\romannumeral2} -- Sensing Paradigms and D\&Is}
\label{fig:taxonomy-system-2}
\end{figure*}

\section{Taxonomy System \expandafter{\romannumeral2}: Classification of mHealth Sensing by Sensing Paradigms and D\&Is}

With respect to distinguished sensing objectives (i.e., \emph{Personalized Medicine} and \emph{Population Health}) and their details discussed in Section \ref{section:objectives}, two sensing paradigms (i.e., \emph{Personal Sensing} and \emph{Crowd Sensing}) are correspondingly proposed to deal with related technical issues through detailed D\&Is. In this phrase, as shown in Figure \ref{fig:taxonomy-system-2}, for each step of mHealth Sensing life cycle, varying detailed D\&I issues of the two sensing paradigms are discussed.

\subsection{Design \& Implementation Issues in Sensing Task Creation \& Participation}

To prompt the users' participation and task execution leveraging \emph{services} and \emph{recruitment} respectively for \emph{Personal Sensing for Personalized Medicine} and \emph{Crowd Sensing for Population Health} apps discussed in \ref{section:ob-participation}, in this section, we intend to specify the detail D\&I issues of the two paradigms as followings.

\paragraph{Personal Sensing for Personalized Medicine}
The promotion of user engagement in Personalized Medicine apps is by providing \emph{services}. Here we discuss two typical forms of user engagement services -- \emph{clinical health service and human-computer interaction (HCI)} and \emph{gamification and attraction} in detail.

\begin{itemize}

    \item \textbf{Clinical Health Service and HCI -} Providing straight-forward and effective clinical health service with good HCI design for user experience is the most intuitive way to increase users' active engagement, since the essential motivation of the users downloading the app is to obtain personal health benefit \cite{poole2013hci,schnall2016user}. In practice, user engagement strategies can be organized as setting sensing health-related targets around users' personalized objectives, delivering adaptive therapeutic feedback including positive reinforcement, reflection reminders, and challenging negative thoughts \cite{steele2009elderly,meng2018exploring}, and designing easy-to-use platforms \cite{boukhechba2020swear}. For instance, Cai \emph{et al.} \cite{cai2021designing,cai2021framework} propose to prompt an adaptive and passive personal mobile sensing framework to provide ecological momentary assessment and intervention services based on the reinforcement learning techniques, which significantly increased user engagement in healthcare apps.

    \item \textbf{Gamification and Attraction -} Gamifying the mHealth Sensing apps for providing entertainment would promote user engagement, as not only the mobile sensing data can be used as input for gamification \cite{fitz2011exploring}, but also mobile apps are excellent and prevailing mobile carriers for pervasive entertainment \cite{lee2017motivates}. In practice, gamification strategies are widely applied in \emph{Personalized Medicine} apps to promote participation such as self-report data collection \cite{crowley2012gamification,rabbi2017sara} (e.g., setting the goals of the game as the indices to be sensed), data pre-analysis on client \cite{l2017gamification} (e.g., pop-up windows asking the user about the activity and status when the app detects a sequence of abnormal indices), and health intervention wrapping \cite{floryan2020model} (e.g., relaxing users under depression via games). Typically, \hl{Rabbi \mbox{\emph{et al.}}{~\cite{rabbi2017sara}} designed an app named SARA, which integrates gamified engagement strategies including contingent rewards, badges for completing active health tasks, funny memes/gifs \& life-insights, and health-related reminders or notifications.}
\end{itemize}

\paragraph{Crowd Sensing for Population Health}

Though participants in \emph{Population Health} tasks may also actively/voluntarily engage in the tasks attracted by D\&Is for services (i.e., \emph{services and HCI designs}) above \cite{agrawal2018towards}, a crucial problem in the tasks does exist -- \emph{participants may not obtain straightforward health benefit compensating their efforts}, leading to its unique incentive mechanisms -- \emph{recruitment with monetary incentives} \cite{zhang2015incentives}. Worth mentioning, in most Crowd Sensing for Population Health practices, the incentive mechanisms (i.e., services and recruitment) are not used strictly separated; they can be wrapped together to optimize the incentive effects \cite{kumar2013mobile,wagner2017wrapper,li2020micro}. 

\begin{itemize}
    \item \textbf{Recruitment with Monetary Incentives -} Monetary incentivization is an intuitive way to quantify and equalize participants' efforts and benefits, though some voluntary Crowd Sensing activities also do exist. \hl{In practice, for research or business purposes, mHealth professionals and insurance companies may consider to promote mHealth apps as tools for groups of interests{~\cite{jo2019there}}.} The monetary incentives strategies can be further divided into categories as platform-centric and user-centric methods \cite{jaimes2017incentivization}. The platform-centric methods refer to that the allocation and adjustment of incentives are charged by the organizers. For example, based on the game theory \cite{dasari2020game}, the organizers can lead the task and adjust the strategies by measuring the individual/overall performance of the participants \cite{yang2012crowdsourcing}. The user-centric methods are mostly conducted in an auction manner, where users bid for the tasks published and the participants with the lowest bid are dynamically allocated to complete the sensing tasks \cite{ji2020reverse}.
\end{itemize}	
\hl{In addition to the above incentive models, there are some works focusing on the participant selection and incentive allocation problems~\cite{wang2013effsense,xiong2017near,xiong2014crowdrecruiter,xiong2014emc,xiong2015crowdtasker,xiong2015icrowd,wang2016fine,wang2018multi} under certain budgets and data collection objectives/constraints, since sometimes too straightforward incentive allocation may lead to biased selection and low retention rate in recruited populations{~\cite{chan2017asthma}}. Specifically, Xiong \emph{et al.} proposed several participant recruitment strategies~\cite{xiong2014crowdrecruiter,xiong2014emc,xiong2015crowdtasker,xiong2015icrowd} for mobile crowd sensing in either online or offline manners. Wang \emph{et al.}~\cite{wang2016fine,wang2018multi} studied the problem of participant recruitment and task/incentive allocation in the context of multi-tasking, where incentives are allocated to the same pool of potential participants for multiple tasks with shared budgets, via hierarchical data collection objectives. The same group of researchers also studied to collect population health-related data from large crowds with non-monetary incentives in practice~\cite{wang2020will,chen2021enabling}.}

\subsection{Design \& Implementation Issues in Health Surveillance \& Data Collection} \label{section:di-data-collection}

\begin{figure}[]
\centering
\includegraphics[width=0.45\textwidth]{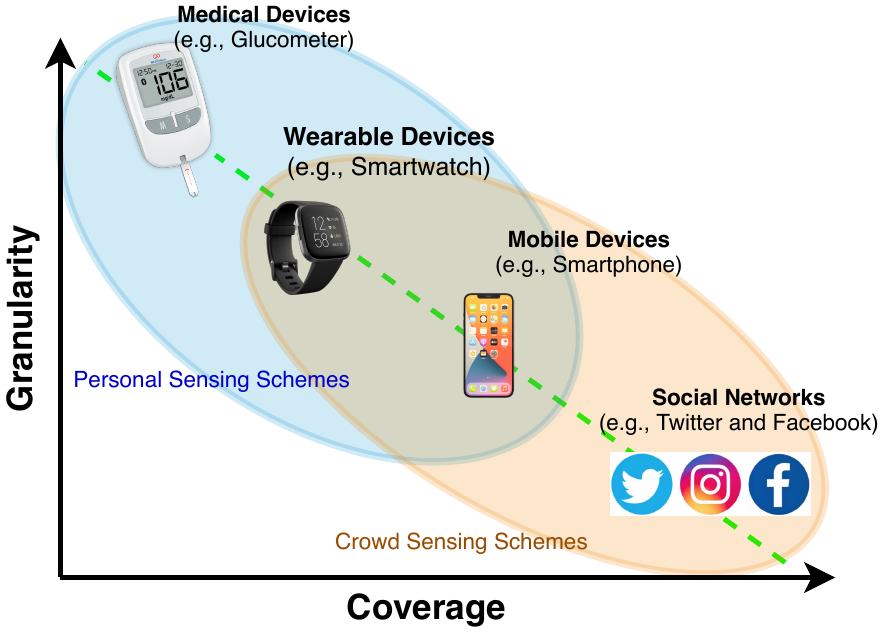}
\caption{The comparison of sensing schemes from granularity and coverage perspectives}
\label{fig:di-schemes}
\end{figure}

In the Health Surveillance \& Data Collection fold, for the objectives of \emph{data quality}, sensing schemes and data gathering approaches are the main D\&I issues. As shown in Figure \ref{fig:di-schemes}, in mHealth Sensing field, either it needs widespread devices (e.g., mobile devices and social network) with pervasive coverage among populations, or it needs dedicated devices (e.g., portable medical devices) for accuracy and professionalism, which is hard to be traded off, limited by the costs and the accessibility of specific devices. Besides, there are some trails, surveys, and evaluations approaches in Crowd Sensing paradigm especially.

\paragraph{Personal Sensing for Personalized Medicine} Though the two paradigms sometimes adopt common sensing schemes (e.g., wearable devices and mobile devices shown in Figure \ref{fig:di-schemes}) under some circumstances, while, for the objectives on numerical accuracy and longitudinal coverage, the sensing schemes in Personal Sensing practices are more granularity-oriented.

\begin{itemize}
\item \textbf{Granularity-Oriented Sensing Schemes -} To accurately monitor user's physical/environmental dynamics in a timely manner, some dedicated and intensive sensors deployed in medical devices are commonly used in Personal Sensing practices, such as mobile fall detection devices on elderly care in daily scenarios \cite{dai2010perfalld,lin2014managing,de2017mobile} and intensive location/maneuvers monitoring devices in hospital scenarios \cite{shirehjini2012equipment,verceles2015use} which are equipped with radar. For example, Fang \emph{et al.} \cite{fang2016bodyscan,fang2016headscan} purposely embedded radio sensor into wearable devices as a new powerful sensing modality to provide whole-body activity and vital sign monitoring in clinical, which serves as an example that specialized sensing schemes provide richer function in Personal Sensing scenarios.
\end{itemize}

\paragraph{Crowd Sensing for Population Health} To broadly collect health-related data with guarantees of population coverage and data fidelity, in Crowd Sensing practice, the detail D\&Is lie on \emph{coverage-oriented sensing schemes} (for population coverage), \emph{trials, surveys, and evaluations} (for data collection efficiency and fidelity).

\begin{itemize}
\item \textbf{Coverage-Oriented Sensing Schemes -} In Crowd Sensing practices, though many sensing schemes are the same as those used in the Personal Sensing apps as shown in Figure \ref{fig:di-schemes}, while, in order to enable the system to be used in a larger population coverage, ubiquitous sensing schemes are prevailing in Crowd Sensing practices, such as social medias (e.g., Facebook and Twitter) \cite{de2013predicting,saha2019social} and large-scale human mobility data which is not gathered dedicatedly for health-related purposes \cite{canzian2015trajectories,boukhechba2017monitoring,wang2020demand}. For instance, Choudhury \emph{et al.} used passive sensed data from social medias to measure and predict the depression in population \cite{de2013social,de2013predicting}, even further to discover shifts to suicidal tendency from content in Reddit \cite{de2016discovering}.

\item \textbf{Trails, Surveys, and Evaluations -} In Crowd Sensing data collection process, it is essential to motivate participants to keep uploading sensing data with efficiency and fidelity. Typically, trail and survey schemes are for the efficiency, and data evaluation schemes are for the fidelity. As for \emph{trails and surveys}, micro-randomized trials (MRTs) are tools for maintaining and improving participants' efficiency by optimizing the combinations of incentives (e.g., varying levels of monetary incentives, and virtual rewards) \cite{walton2018optimizing,seewald2019practical,li2020micro}. With MRTs, participants first randomly grouping to collect data under varying incentives, then in the following sensing loops, the collected data in the previous round is used to measure which combinations of incentives are optimal. As for \emph{evaluation schemes}, they are for enforcing data fidelity \cite{restuccia2017quality}. In specific, once a new round of data \emph{collected}, but before accepting the data as convinced, the data fidelity is estimated and only convinced data is \emph{gathered}; according to the estimation, positive or negative feedback is given to participants to reward/punish them in the following rounds. An intuitive scheme, named truth discovery \cite{li2016survey}, is to let multiple participants finish a same task to find the wrong-data providers \cite{meng2015truth}. However, this repeated validation manner cannot be adopted to health-related data collection since sensitive personal data can only be sensed by the individual himself/herself. While the trust framework \cite{mousa2015trust} is an alternative means to solve this. Some measurement methods can be used to establish a credit rating measurement system for participants, and implement different acceptance of data contributed by users with different credits, and varied tasks and incentives are dynamically allocated to enforce participants' performance in the following sensing rounds \cite{jin2015quality,guo2017taskme,jiang2020toward}.
\end{itemize}

\begin{figure*}[]
\centering
\includegraphics[width=0.8\textwidth]{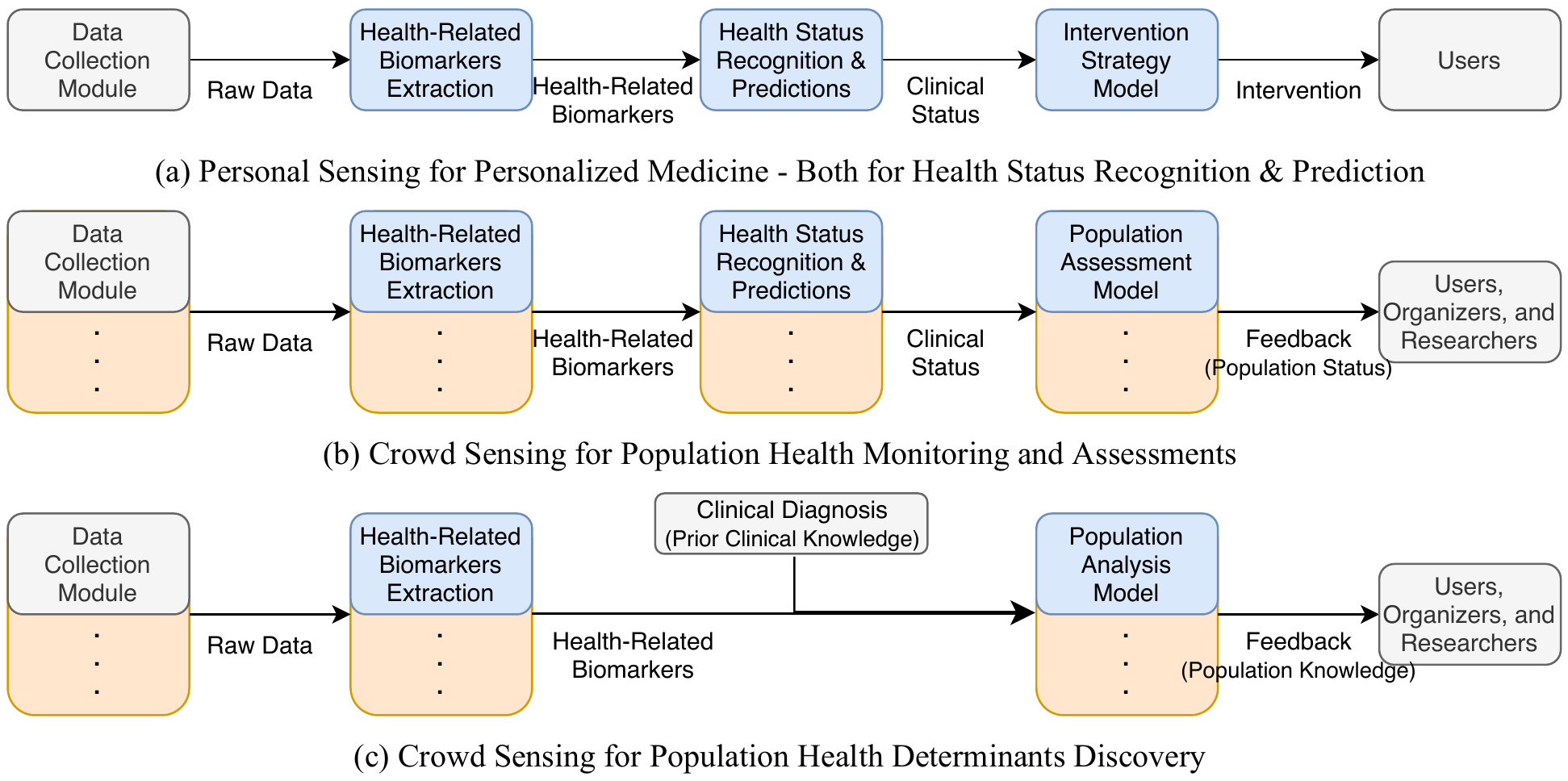}
\caption{Data Analysis \& Knowledge Discovery workflows in three typical Personal Sensing and Crowd Sensing apps.}
\label{fig:di-data-analysis}
\end{figure*}

\subsection{Design \& Implementation Issues in Data Analysis \& Knowledge Discovery}

With respect to detailed sensing objectives listed in Section \ref{section:ob-data-analysis}, we one-by-one discuss the detail D\&I issues in this section. Furthermore, in order to fully study detailed technical perspectives of the two sensing paradigms, inspired by the mHealth Personal Sensing framework proposed by Mohr \emph{et al.} \cite{mohr2017personal}, we formulate the D\&Is of Data Analysis \& Knowledge Discovery workflow as shown in Figure \ref{fig:di-data-analysis}.

\paragraph{Personal Sensing for Personalized Medicine} Generally, in Personal Sensing, Data Analysis \& Knowledge Discovery serves to mine collected raw data to realize \emph{health status recognition and interventions} or \emph{health outcomes and progression predictions}.

\begin{itemize}
    \item \textbf{Health Outcomes and Progression Predictions -} Due to the fact that most health problems are determined by multiple pathogenic factors and sometimes progress slowly, it is not trivial for conventional clinical methods to effectively predict health outcomes and progressions via sparse clinical records \cite{elf2017case}. Personal Sensing data provides rich personalized information to model the health status of user and predict his/her future health outcomes and progressions. As shown in Figure \ref{fig:di-data-analysis} (a), after collecting \emph{raw data} (e.g., GPS location, microphone signal, and screen status), digital physical and environmental biomarkers (e.g., places, ambient noises, and app usages) can be extracted \cite{wang2018sensing,buck2019capturing}. Then, personal health status modeling and prediction models analyze individuals' \emph{clinical status} and predict health outcomes and progressions with consideration of longitudinal data both in current and historical. For instance, in the machine learning era, feature embedding and deep learning techniques are good tools to solve the challenges in multidimensional pathogenic factors and long-term disease progression; specifically, feature embedding techniques (e.g., graph embedding) automatically learn and extract influential features \cite{cruciani2020feature}, and deep learning models (e.g., RNNs, GNNs) could serve as predictors with great performance in dynamically capturing patterns in temporal and other dimensions \cite{jimenez2015mobile,boukhechba2018predicting,zhao2019indoor,tseng2020using,dong2021using}.
    
    \item \textbf{Health Status Recognition and Interventions -}  As shown in Figure \ref{fig:di-data-analysis} (a), according to different health status, the Personal Sensing apps could deliver varying interventions as healthcare services for users. What's more, the apps can further recognize users' following status for measurement of the interventions' effectiveness to refine the strategies and suit the users \cite{aung2017sensing,costa2019boostmeup}. \hl{As for implementations, activity recognition approaches are helpful for health status modeling and recognition~\cite{chen2011knowledge,chen2011activity,chen2012sensor}. Okeye \emph{et al.} \cite{okeyo2012knowledge,okeyo2013agent,okeyo2014dynamic} proposed multiple-sensors based activity recognition schemes by extracting knowledge from smart ambiences; and Triboan \emph{et al.} \cite{triboan2017semantic,triboan2017real,triboan2019semantics,triboan2019fuzzy} improved the activity recognition methods to be applied in complex environments in a more real-time and fine-grained manner.} Besides, MRTs \cite{hertel2011cognitive} are ideal tools to deliver JITAI for patients. As stated in Section \ref{section:di-data-collection}, analogous to the designs of MRTs in improving the effectiveness of interventions.

\end{itemize}

\paragraph{Crowd Sensing for Population Health} We discuss two typical applications (i.e., \emph{population health status measurement} and \emph{health determinants discovery}) to conclude D\&I in Crowd Sensing applications.

\begin{itemize}
    \item \textbf{Population Health Monitoring and Assessments -} Intuitively, as shown in Figure \ref{fig:di-data-analysis} (b), once sensing tasks among a group of users are adopted, organizers can scan the clinical status among populations and achieve assessment of population status. Furthermore, in the population assessment models, some techniques (i.e., transfer learning \cite{spathis2021self}) inspired by some characteristics of population health problems, such as spatial correlation, help achieve low-error surveys of entire target group by only monitoring a subset of users. For example, to investigate a large group of people such as citizens of a country, Chen \emph{et al.} \cite{chen2021enabling} studied and indicated spatio-temporal correlation of neighboring regions and proposed to do data inference for the whole map with limited region samples, which gives insights in operating population health monitoring in a Crowd Sensing manner.
    
    \item \textbf{Health Determinants Discovery -} As shown in Figure \ref{fig:di-data-analysis} (c), the D\&Is of applications on population health determinants discovery differ. Specifically, in clinical practices, especially for mental health and chronic illness, with prior knowledge such as clinical diagnosis and EMA, organizers massively collecting multi-modal data from participants (participants may be divided into experimental group and control group) and analyze population pattern among participants' biomarkers and clinical diagnosis to understand and discover health determinants; finally, population knowledge serves as feedback, which benefits to both participants themselves (for health-related interests) and organizers and researchers (for knowledge about the health issues researched). From the implementation perspective, large-scale data analysis methods give great insights on population health knowledge discovery (e.g., inference and understanding) from Crowd Sensing data. For instance, machine learning methods such as clustering algorithms are widely used to classify individuals into groups according to common health-related patterns \cite{delias2015supporting}. In addition, statistical methods such as statistical inference are also promising confirmatory tools for understanding and inference on clinical conclusion than training-based models with confidence intervals on assessment, which, compared with machine-learning-based methods, is commonly leveraged by clinical scientists since it is a hypothesis-driven and interpretable manner \cite{intille2013closing}. For example, Boukhechba \emph{et al.} \cite{boukhechba2017monitoring} used \emph{Social Interaction Anxiety Scale} (SIAS) correlation analysis to understand how social anxiety symptoms manifest in the daily lives of college students; besides correlation analysis, Huang \emph{et al.} \cite{huang2016assessing} operated a \emph{Least Absolute Shrinkage and Selection Operator} (LASSO) linear regression model to infer the causal relationship between mental health disorders and location semantics.
\end{itemize}
Though items above could summarize most of the D\&Is issues in this step in Personal Sensing and Crowd Sensing paradigms, there are also some side D\&Is issues for some problems that may exist in the mobile sensing data \cite{tang2020exploring}. For example, ideally, the input of the data analysis algorithm is continuous and sufficient, while in mHealth Sensing contexts, the data streams collected may be sparse and biased due to some technical issues (e.g., operating system's restrictions on software running in the background) and varying users' usage behaviors (e.g., forgetting to wear the device or run the app); thus overcoming the insufficiency of data and effective modeling is an urgent problem to be solved \cite{sarker2016finding,boukhechba2020leveraging}. \hl{Additionally, similar side problems include ways to analyze and understand the relationship between the complex dynamics of the health and multimodal factors \cite{shoaib2016complex}, and ways to integrate medical knowledge into algorithms pervasively and effectively \cite{huckins2019fusing}.}

\begin{table*}[]
\caption{Summary of two taxonomy systems for mHealth Personal Sensing and Crowd Sensing}
\centering
\begin{tabular}{|c|l|l|l|l|}
\hline
\multirow{2}{*}{
} & \multicolumn{2}{c|}{\textbf{\begin{tabular}[c]{@{}c@{}}Sensing Objectives (System I)\end{tabular}}} & \multicolumn{2}{c|}{\textbf{Sensing Paradigms and D\&Is (System II)}} \\ \cline{2-5} 
 & \multicolumn{1}{c|}{\textbf{\begin{tabular}[c]{@{}c@{}}Personalized Medicine\end{tabular}}} & \multicolumn{1}{c|}{\textbf{Population Health}} & \multicolumn{1}{c|}{\textbf{Personal Sensing}} & \multicolumn{1}{c|}{\textbf{Crowd Sensing}} \\ \hline
 
\textbf{\begin{tabular}[c]{@{}c@{}}Sensing\\ Task Creation \\ \& \\ Participation\end{tabular}} & \begin{tabular}[c]{@{}l@{}}Service provision \\for personalized \\medicine seekers \end{tabular} & \begin{tabular}[c]{@{}l@{}}Recruitment provision \\for population \\health participants\end{tabular} & \begin{tabular}[c]{@{}l@{}} Improving user engagement \\via clinical health services, \\gamification and attractions \end{tabular} & \begin{tabular}[c]{@{}l@{}}Extra motivating participants' \\performance via recruitment \\with monetary incentives\end{tabular} \\ \hline

\multirow{2}{*}{\textbf{\begin{tabular}[c]{@{}c@{}}Health\\ Surveillance \\ \&\\ Data  Collection\end{tabular}}} & \begin{tabular}[c]{@{}l@{}}More focusing on data \\precision and longitudinal \\ coverage\end{tabular} & \begin{tabular}[c]{@{}l@{}}More focusing on data\\ fidelity and population\\ coverage\end{tabular} & \multirow{2}{*}{\begin{tabular}[c]{@{}l@{}}Using granularity-oriented\\ sensing schemes\end{tabular}} & \multirow{2}{*}{\begin{tabular}[c]{@{}l@{}}Using coverage-oriented\\ sensing schemes with trails,\\ surveys, and evaluations\end{tabular}} \\ \cline{2-3}
 & \multicolumn{2}{l|}{\hl{Privacy \& Security and Resources Consumption}} &  &  \\ \hline

\multirow{2}{*}{\textbf{\begin{tabular}[c]{@{}c@{}}Data Analysis \\ \& \\ Knowledge\\ Discovery\end{tabular}}} & \begin{tabular}[c]{@{}l@{}}Improving mHealth\\ accuracy in risk \\prediction and precision \\in predictive interventions\end{tabular} & \begin{tabular}[c]{@{}l@{}}Pursuing depth and \\coverage of health \\monitoring in communities\\ and statistical power \\of mHealth approaches\end{tabular} & \multirow{2}{*}{\begin{tabular}[c]{@{}l@{}}Leveraging health outcomes \\and progressions predictions,\\ and health status recognition \\and interventions\end{tabular}} & \multirow{2}{*}{\begin{tabular}[c]{@{}l@{}}Leveraging population health\\ assessment and health \\determinants discovery\end{tabular}} \\ \cline{2-3}
 & \multicolumn{2}{l|}{\hl{Risks and Ethical Issues}} &  &  \\ \hline

\end{tabular}
\label{taxonomy-system}
\end{table*}

\section{Future Directions}
\hl{In this work, we reviewed the applications and systems of personal sensing and crowd sensing for personalized medicine and population health, respectively, and proposed two taxonomy systems for mHealth Sensing systems from the perspectives of \emph{``Sensing Objectives''} and \emph{``Sensing Paradigms'' }. Here summarize the two taxonomy systems in Table~\ref{taxonomy-system}.}
\hl{It is obvious that mHealth Sensing apps in both \emph{Personal Sensing} and \emph{Crowd Sensing} paradigms will continue to be promising research topics to solve both \emph{Personalized Medicine} and \emph{Population Health} problems, where some research problems such as data limitations, data fidelity, privacy \& security, risk analysis, and ethical issues are still not well addressed in mHealth Sensing life cycle. Based on the proposed taxonomy systems and identified gaps, we foresee the following research directions in future works.}

\hl{{\subsection{Data Limitation and Data Fidelity}}
In mHealth contexts, the potential future directions in terms of data could be solving the research problems of data limitations and data fidelity. First, the \emph{data limitations} in time series, as well as between sensor samplings and system operations~\cite{kayacik2014data,badr2021limitations} lead to discontinuous multimodal data collection and even loss in mHealth apps~\cite{wan2019multi}. Existing mHealth data analysis works lack design for handling imperfect heterogeneous data, where transfer learning techniques may be potential tools \cite{spathis2021self}. Second, the \emph{data fidelity} issues caused by participants' concealment or deception lead to biased/error data gathering and misleading/false health conclusions~\cite{schweitzer2012economics}. Thus, besides of promoting user engagement by incentive strategies, the work of effectively verifying the fidelity of data uploaded by users is worthy of further study.
}

\hl{{\subsection{Privacy and Security Preserving for mHealth}}
Note that privacy and security have been widely studied in Medical IoTs or Medical CPSs~\cite{gayathri2019efficient,atat2018a,singh2018secure,parah2020efficient,peng2020secure,liu2020improved,kumar2020a}. Compared to medical IoTs or medical CPSs deployed at homes or professional clinics, the mHealth sensing systems leveraging the sensors deployed at ubiquitous mobile devices make the privacy and security issues even more complicated but lack of studied. To secure the personal health data from potential leakages, encryption techniques~\cite{ibraimi2009secure,tang2019efficient} could be used and optimized for mHealth data management. Additionally, privacy protection that controls the access of mobile Apps to some critical information~\cite{jain2012addressing,zhu2014mobile} is also required to scale-up mHealth in societies. In this way, mobile developers frequently need to design and develop the apps with verification that minimizes data access privileges subject to the actual needs. Thus, a unified and integrated approach, combining the data security and privacy controls subject to \emph{principle of least privilege}~\cite{sandhu1996role,felt2011android,roesner2012user} for mHealth sensing, might be a promising direction for future research.}

\hl{{\subsection{Risks and Ethical Issues in Human-Subject Studies}}
After-all, the research on mHealth sensing is human-subject studies, where human involve in-the-loop of scientific studies, data analysis, and information disclosures, causing potential risks and ethical issues. Though some works have been done in software developing and data science domains~\cite{lurie2016professional,mittelstadt2019principles}, versus clinical medical practices which pay great attention to risk analysis and ethical principles~\cite{carden2021ethical}, the risk analysis and ethical issues in mHealth area are not properly studied and addressed~\cite{mittelstadt2017designing,mittelstadt2017ethics}. For example, versus medical records and conclusions are drawn under highly professional processes and stored separately by hospitals' databases with strict rules for sharing, the measurements and decisions in mHealth practices may not be strictly conducted and shared under criterion~\cite{fiordelli2013mapping}. Truly, some of mHealth sensing apps, such as Sensus~\cite{xiong2016sensus}, already include protocol certification and ethical review components in the system to monitor the whole life-cycles of mHealth crowd sensing. In the future, scientific study, protocol management, risk analysis, ethical review, and even prescription management~\cite{hossain2018an} criteria and techniques should be further studied, especially for commercially-used mHealth sensing apps and systems.
}

\section{Limitations and Conclusions}
The mHealth Sensing is a practical approach in modern healthcare domain, which is being widely used for the objectives on either \emph{(a) personalized medicine for individuals} or \emph{(b) public health for populations}. In this work, we reviewed and summarized mHealth sensing Apps and systems that deployed over smartphones and commodity ubiquitous devices. \hl{Though there are many methods for reporting systematic reviews (e.g., PRISMA \cite{page2021prisma}), in this paper, our review method is mainly intuition-driven and vision-based. We have covered more than 300 papers and proposing new taxonomy systems that summarize and categorize existing works in two sensing paradigms (i.e., Personal and Crowd Sensing) and three stages of the mHealth sensing pipeline in details.
}

Also, though we have tried our best to cover the important works in this area and related fields, this survey is still with several limitations. For example, this work did not include professional medical systems for medicare/rehabilitation/assisted living purposes, such as medical sensors~\cite{beck2000remote,chen2014continuous,khakh2006p2x,guntner2019breath,gao2020flexible,senesky2009harsh}, Medical IoTs/CPSs~\cite{gatouillat2018internet,dimitrov2016medical,haghi2017wearable,hayani2020image,joyia2017internet,elhoseny2019effective,pazienza2020adaptive,limaye2018hermit,gayathri2019efficient,atat2018a,haghi2020a,rahman2018m,jiang2019toward,parah2020efficient,peng2020secure,liu2020improved,zhang2019energy,meng2021hybrid,amendola2014rfid,lin2020covid,sundaravadivel2018everything,hossain2018an,zhu2015bridging}, and medical robots~\cite{altaee2017robot,taylor2003medical,webster2006nonholonomic,wang2011lower,beasley2012medical,fukushima2014medical}. Furthermore, there have been a number of great works surveying or reviewing this area and related fields~\cite{islam2015the,santos2020online,mittelstadt2017ethics,gatouillat2018internet,banaee2013data,sun2018security,zhou2010case,martinez2015privacy,jothi2015data,verceles2015use,silva2015mobile,martinez2013mobile,trifan2019passive,kalantarian2017survey,wang2020privacy,restuccia2017quality,pournajaf2014survey,zhang2015incentives,lane2010survey,dasari2020game,surantha2016internet}, while we have not compared our taxonomy systems with these works.

To systematically summarize the existing works and identify the potential directions in this emerging research domain, this work actually presents two novel taxonomy systems from two major perspectives (i.e., \emph{sensing objectives} and \emph{sensing paradigms and Designs \& Implementations (D\&Is)}) that can specify and classify apps/systems from steps in the life-cycle of mHealth Sensing: \emph{(1) Sensing Task Creation \& Participation}, \emph{(2) Health Surveillance \& Data Collection}, and \emph{(3) Data Analysis \& Knowledge Discovery}. Through discussing the real-world Mobile Sensing apps/systems in the proposed taxonomy systems, most of the research problems in mHealth Sensing can be formally classified, and several future research directions are pointed out, targeting to provide structural knowledge and insightful ideas and guidance for researchers in the related field.

\bibliographystyle{IEEEtran}
\bibliography{main}
\clearpage
\appendix
\section*{Discussion on Scientific Approach}
\hl{In the appendix, we discuss the scientific approach of this survey. First of all, we would like to clarify our motivation --- mHealth sensing, where we include a brief discussion on the comparisons between mHealth sensing and sensing techniques in general medical settings. Later, we review the scientific procedures and criteria that we select publications for review. Finally, we review the scientific way that we built the two taxonomy systems.}

\hl{\subsection{Research Definition: mHealth Sensing versus Medical Sensing}}

\hl{In this paper, we give a comprehensive survey on \emph{mHealth Sensing} techniques, where the topic (e.g., mHealth Sensing) is close but significantly from ``\emph{Medical Sensing}''. Here we discuss the major differences between the two types of sensing techniques in a structured manner, including ``target populations'' who need the two sensing techniques, ``deployment contexts'' that the two sensing techniques are adopted, ``medical goals'' that the two sensing techniques aim to meet, and ``methodologies'' that the two sensing techniques propose to collect data.}

\hl{As shown in Table {\ref{sensing-versus}}, the \emph{mHealth Sensing} and \emph{Medical Sensing} techniques vary significantly. Specifically, \emph{mHealth Sensing} techniques are majorly designed for voluntary/active users in some daily/commercial scenarios, for health-related behaviors monitoring and intervention, by leveraging wearable/mobile devices. The overall goals of mHealth Sensing are improving health status and well-beings by studying the innovative applications of mobile sensing techniques to collect behavioral and environmental data. In contrast, the \emph{Medical Sensing} takes care of patients in medicare context such as hospital or clinical scenarios, where it aims at providing the medical diagnoses and treatments through professional devices, where the overall goals of Medical Sensing is to design new measurement and instrument techniques for medical purposes. Thus, the mHealth Sensing mainly discussed in this paper essentially differ from Medical Sensing.
}

\begin{table}[!htbp]
\caption{\hl{mHealth Sensing versus Medical Sensing}}
\centering
\hl{
\begin{tabular}{|l|c|c|}
\hline
                  & \textbf{mHealth Sensing}    & \textbf{Medical Sensing}   \\ \hline
\textbf{Populations}      & Voluntary or active users   & Patients in medical care   \\ \hline
\textbf{Contexts}    & Daily or commercial         & Hospital or clinical       \\ \hline
\textbf{Coverage} & Monitoring and intervention & Diagnosis and treatment    \\ \hline
\textbf{Goals}     & Health issues for well-being & Medical issues for clinics \\ \hline
\textbf{Methods}      & Wearable/mobile devices     & Professional devices       \\ \hline
\end{tabular}
}
\label{sensing-versus}
\end{table}

\hl{\subsection{Research Methods: Selection Procedure and Criteria for Related Works}}

\hl{In tis survey, we refer more than 300 technical papers and review them from applications and taxonomies perspectives. Generally, we collect these publications in scientific way as follows.}

\begin{itemize}

    \item \hl{First, we cover several notable publications from our previous works in \emph{mHealth Sensing}, mobile systems design, and crowd sensing and data analytics. Specifically, we include our works on mobile systems design to understand the dynamics and personalization of health and well-being \cite{xiong2015icrowd,huang2016assessing,xiong2017near,chow2017using,boukhechba2017monitoring,boukhechba2018predicting,boukhechba2020leveraging,boukhechba2020swear,cai2021framework,cai2021designing}. We also introduce our works in smart healthcare, elderly care and context-aware computing for health~\cite{du2008hycare,zhou2010case,lin2012detecting,zhang20144w1h,xiong2015icrowd,wang2016fine,guo2017taskme,wang2018multi,wang2018energy,wang2020will}, and the fundamentals of mobile crowd sensing and crow data analysis for health \cite{xiong2014crowdrecruiter,xiong2014emc,xiong2015eemc,xiong2015icrowd,xiong2015crowdtasker,wang2016fine,huang2016assessing,xiong2016sensus,chow2017using,xiong2017near,wang2018multi,huang2020quantifying,liu2021analysis,Xiong2020.04.20.20068676}. The inclusion of our previous works further confirms the professionality of our group to carry out this survey. }

    \item \hl{Besides, we also cover some notable works from pioneering research groups in mHealth sensing domains. Specifically, we cover works from Campbell's lab that focus on developing mobile sensing technology capable of accessing mental health \cite{lane2010survey,lane2011bewell,wang2016crosscheck,harari2016using,wang2017predicting,wang2018sensing,saha2019social,buck2019capturing,tseng2020using,ben2020mobile}, Murphy's works on decision making and interventions in mHealth \cite{wagner2017wrapper,walton2018optimizing,nahum2018just,menictas2019artificial,seewald2019practical,bidargaddi2020designing,menictas2020fast,li2020micro}, Choudhury's works on the mHealth sensing systems for people's context, activities and social networks \cite{choudhury2008mobile,lane2011bewell,rabbi2011passive,chen2013unobtrusive,abdullah2014towards,rabbi2015automated,wang2016crosscheck,costa2017emotioncheck,wang2017predicting,aung2017sensing,abdullah2018sensing,rey2018personalized,costa2019boostmeup,buck2019capturing,wang2020social,tseng2021digital}, as well as Mascolo's works on mobile and wearable systems design to human behaviour understanding from mHealth perspectives \cite{servia2017mobile,tang2020exploring,brown2020exploring,spathis2021self,perez2021wearables,han2021exploring}. Of-course, we also include and discuss works from other important groups.}

    \item \hl{Later, publications are selected from leading technical conferences and journals in the related areas, such as IEEE Internet of Things Journal \cite{wang2016fine,wu2017dynamic,anjomshoaa2018city,ji2020reverse,liu2018large,xu2021beeptrace}, IEEE Communications Magazine \cite{lane2010survey,ganti2011mobile,zhang20144w1h,xiong2017near,wang2018energy}, IEEE Pervasive Computing \cite{intille2013closing,choudhury2008mobile}, \hl{IEEE Sensors} \cite{shoaib2016complex,jovanovic2019mobile,majumder2019smartphone,dasari2020game}, IEEE Transactions on Mobile Computing \cite{xiong2014emc,xiong2015icrowd,wang2018multi}, Proceedings of the ACM Transactions on Interactive, Mobile, Wearable and Ubiquitous Technologies (aka., ACM UbiComp) \cite{wang2013effsense,abdullah2014towards,xiong2014crowdrecruiter,canzian2015trajectories,huang2016assessing,wang2016crosscheck,xiong2016sensus,rabbi2017sara,wang2017predicting,wagner2017wrapper,boukhechba2017monitoring,wang2018sensing,wang2020will,sharmila2020towards,zhang2020necksense,zhang2021passive}, as well some medical journals such as JAMA \cite{steinhubl2013can,mega2014population,bell2020frequency,gao2020association}, Journal of Medical Internet Research \cite{burns2011harnessing,saeb2015mobile,karhula2015telemonitoring,asselbergs2016mobile,chow2017using,boonstra2018using,meng2018exploring,pryss2020applying,floryan2020model,kalanadhabhatta2021effect}, New England Journal of Medicine \cite{seaquist1989familial,manary2013patient}, the Lancet \cite{koehler2018efficacy,zhou2020effects,badr2021limitations}, Nature and Nature Communication \cite{anthes2016mental,wood2019taking,ienca2020on,grantz2020the,tseng2021digital}, Frontiers in neuroscience \cite{huckins2019fusing,kraft2020combining}. Of-course, we also include and discuss works from other important journals and conferences.}

    \item \hl{Finally, we also cover the publications contributed by the related projects funded by National Science Foundation (NSF), National Institute of Health (NIH) and other funding agency. Actually, we pay special attentions to the works from Mobile Sensor Data to Knowledge (MD2K)~\cite{Kumar2015}\footnote{\url{https://md2k.org/research-agenda/pubs.html}} which was established by the National Institutes of Health Big Data to Knowledge Initiative, and also the Trustworthy Health \& Wellness (THaW)\footnote{\url{https://thaw.org/}, \url{https://www.zotero.org/groups/2647330/thaw/library}} that aims at ``Making mobile health effective and secure''~\footnote{\url{https://www.nsf.gov/discoveries/disc_summ.jsp?cntn_id=137188}}}.
    
\end{itemize}
\hl{Based on the above procedures, we collect, review and discuss related works in mHealth sensing areas.}

\hl{\subsection{Scientific Methods in Taxonomy Systems}}
\hl{In this paper, we followed a ``Descriptive and Mapping Reviews''~\footnote{Chapter 9 Methods for Literature Reviews, Handbook of eHealth Evaluation: An Evidence-based Approach [Internet], \url{https://www.ncbi.nlm.nih.gov/books/NBK481583/}} pattern to organize the survey, where we extracted a body of knowledge from existing research works on mHealth sensing. The body of knowledge included a list of health issues related to mHealth sensing and the publications, two taxonomy systems summarizing and classifying the research topics in mHealth sensing area.} 

\hl{Actually, we first reviewed the existing mHealth sensing apps from the perspectives of health issues, where a particular attention to the significant health issues that have been widely studied in mHealth sensing has been paid. We summarized and generated the list of health issues that we took care of from the collected publications (see also in Appendix B). We also discussed the important health issues, such as addictive behaviors, that we might ignore in this survey.} 

\hl{For the two taxonomy systems, we followed simple ``Narrative Reviews''~\footnote{Samuel J. Stratton, MD, MPH, Literature Reviews: Methods and Applications, \url{https://www.cambridge.org/core/services/aop-cambridge-core/content/view/70581E0B68B491693E8360DE39E0D6E4/S1049023X19004588a.pdf/literature_reviews_methods_and_applications.pdf}} strategies on classifying literature works: (1) what technical problems they aimed to solve in the paper (objectives), and (2) how they solved the problem (methodologies). With these two strategies in mind, we develop the taxonomy systems on ``mHealth Sensing Objectives'' and ``D\&I Issues of mHealth Sensing'' that classify the existing works according to their sensing problems and solutions respectively. Of-course the interplays between objectives and D\&I issues are also discussed in an ad-hoc manner. After-all, we demonstrate our visions in the area.}

%

%
\end{document}